\documentclass[lettersize,journal]{IEEEtran}
\usepackage{graphicx}
\usepackage{booktabs} % For formal tables
\usepackage{times}  %Required
\usepackage{helvet}  %Required
\usepackage{courier}  %Required
\usepackage{url}  %Required
\usepackage{graphicx}  %Required
\usepackage{color}
\usepackage{amsfonts,amssymb}
\usepackage{amsmath}
\usepackage[ruled,linesnumbered,noend]{algorithm2e}
\usepackage{diagbox}
\SetKwProg{Fn}{Function}{}{end}
\SetKwFunction{MPCACFL}{MPCAC-FL}
\SetKwFunction{CACFL}{CAC-FL}
\usepackage{subfigure}
\usepackage{multirow}

\newcommand{\eg}{\emph{e.g.},\xspace}
\newcommand{\ie}{\emph{i.e.},\xspace}
\newcommand{\etc}{etc.\xspace}
\newcommand{\new}[1]{{\color{black}{#1}}}

\newtheorem{problem}{Problem}
\newtheorem{definition}{Definition}

\begin{document}

\title{Convergence-aware Clustered Federated Graph Learning Framework for Collaborative Inter-company Labor Market Forecasting}

\author{
    Zhuoning~Guo,
    Hao~Liu,~\IEEEmembership{Senior~Member,~IEEE},
    Le~Zhang,
    Qi~Zhang,
    Hengshu~Zhu,~\IEEEmembership{Senior~Member,~IEEE},
    Hui~Xiong,~\IEEEmembership{Fellow,~IEEE}
    \IEEEcompsocitemizethanks{
        \IEEEcompsocthanksitem Zhuoning Guo is with the Thrust of Artificial Intelligence, The Hong Kong University of Science and Technology (Guangzhou), Guangzhou, China.\\
        E-mail: zguo772@connect.hkust-gz.edu.cn
        \IEEEcompsocthanksitem Hao Liu is with the Thrust of Artificial Intelligence, The Hong Kong University of Science and Technology (Guangzhou), Guangzhou, China and Guangzhou HKUST Fok Ying Tung Research Institute, Guangzhou, China.\\
        E-mail: liuh@ust.hk
        \IEEEcompsocthanksitem Le Zhang is with Baidu Research, Baidu Inc., 100085, Beijing, China.\\
        E-mail: zhangle0202@gmail.com
        \IEEEcompsocthanksitem Qi Zhang is with Shanghai Artificial Intelligence Laboratory, Shanghai, China\\
        E-mail: zhangqi.fqz@gmail.com
        \IEEEcompsocthanksitem Hengshu Zhu is with Career Science Lab, BOSS Zhipin, 100028, Beijing, China.\\
        E-mail: zhuhengshu@gmail.com
        \IEEEcompsocthanksitem Hui Xiong is with the Thrust of Artificial Intelligence, The Hong Kong University of Science and Technology (Guangzhou), Guangzhou, China and Guangzhou HKUST Fok Ying Tung Research Institute, Guangzhou, China.\\
        E-mail: xionghui@ust.hk
        \IEEEcompsocthanksitem Hao Liu is the corresponding author.
    }
}

\markboth{Journal of \LaTeX\ Class Files,~Vol.~14, No.~8, August~2021}%
{Shell \MakeLowercase{\textit{et al.}}: A Sample Article Using IEEEtran.cls for IEEE Journals}


\maketitle

\begin{abstract}
\new{
Labor market forecasting on talent demand and supply is essential for business management and economic development.
With accurate and timely forecasts, employers can adapt their recruitment strategies to align with the evolving labor market, and employees can have proactive career path planning according to future demand and supply.
However, previous studies ignore the interconnection between demand-supply sequences among different companies and positions for predicting variations.
Moreover, companies are reluctant to share their private human resource data for global labor market analysis due to concerns over jeopardizing competitive advantage, security threats, and potential ethical or legal violations.
To this end, in this paper, we formulate the Federated Labor Market Forecasting~(FedLMF) problem and propose a Meta-personalized Convergence-aware Clustered Federated Learning~(MPCAC-FL) framework to provide accurate and timely collaborative talent demand and supply prediction in a privacy-preserving way.
First, we design a graph-based sequential model, combining a Demand-Supply Joint Encoder-Decoder and a Dynamic Company-Position Heterogeneous Graph Convolutional Network to capture the inherent correlation between demand and supply sequences and company-position pairs.
Second, we adopt meta-learning techniques to learn effective initial model parameters that can be shared across companies, allowing personalized models to be optimized for forecasting company-specific demand and supply, even when companies have heterogeneous data.
Third, we devise a Convergence-aware Clustering algorithm to dynamically divide companies into groups according to model similarity and apply federated aggregation in each group. The heterogeneity can be alleviated for more stable convergence and better performance.
Extensive experiments demonstrate that MPCAC-FL outperforms compared baselines on three real-world datasets and achieves over $97\%$ of the state-of-the-art model, \ie DH-GEM, without exposing private company data.
}
\end{abstract}


\section{Introduction}
\IEEEPARstart{T}{alent} acquisition war has intensified in labor markets due to industrial upgrading in recent years~\cite{black2021ai}. Organizations and companies continuously review and adapt their recruitment strategies to align with the radically varied labor market, which raises an urgent need for labor market forecasting. As an essential block of labor market analysis, labor market forecasting aims to model the landscape of the time-evolving labor market, including both talent demand~\cite{zhang2021talent, zhu2016recruitment} and supply~\cite{li2017nemo, zhang2021attentive} variation. Indeed, timely and accurate forecasting of the labor market trend not only helps the government and companies with policy and recruitment strategy readjustment but is also beneficial for job seekers to plan their career path proactively~\cite{cappelli2014talent}.

Extensive studies have been conducted for labor market forecasting encompassing various perspectives.
Conventional heuristic methods are primarily concentrating coarse-grained labor market analysis~(\eg industry-specific demand trend~\cite{lin2011factors} and geographic-occupational labor market concentration~\cite{azar2022labor}) based on survey data~\cite{sia2013university}.
Such methods rely on classical statistical models and domain expertise but fail to account for more complex latent data dependencies. The new emerging data-driven methods utilize machine learning techniques to exploit large-scale data acquired from online professional platforms. For example, TDAN~\cite{zhang2021talent} employs an attention mechanism to forecast talent demand values for upcoming time intervals based on observed data and Ahead~\cite{zhang2021attentive} integrates a Dual-Gated Recurrent Unit model with heterogeneous graph embeddings to predict the next moving company, position, and working duration from the supply perspective.
These methods treat talent demand or supply forecasting as a time series prediction task, where various sequential deep learning models have been introduced to capture the underlying temporal correlations of market trend variations.

After analyzing large-scale real-world data, we identify two important labor market variation characteristics, which have been rarely considered in previous studies. On the one hand, talent demand and supply are intrinsically correlated with each other.
For example, the emerging demand of a rising company will attract more talents, and the oversupply of a position may curb the demand for an extended period to resolve the excessive talents.
Modeling the interconnection between talent demand and supply variation can provide extra information for both tasks to predict more precisely.
On the other hand, the demand-supply variation of different companies and positions are correlated yet diversified.
Companies and positions in the same industry may follow similar co-evolvement patterns~\cite{cappelli2008talent,aviv2001effect}, \eg BYD and Tesla may recruit many computer vision engineers in the trend of self-driving.
However, even subsidiaries of the same company may have very different talent demand requirements at different times.
Distilling and incorporating the shared knowledge between related companies and positions can further improve the effectiveness of both talent demand and supply forecasting.
Inspired by the above characteristics, in this work, we study the talent demand-supply joint prediction problem, where the talent demand and supply of positions in every company are predicted simultaneously.

Three major challenges arise toward talent demand-supply joint prediction.
First, existing labor market forecasting methods either focus on talent demand or supply prediction but overlook the intrinsic correlation between talent demand and supply variation. It is challenging to incorporate the interconnection between two different tasks in a mutually reinforcing way.
Second, the correlation between different companies and positions may vary. Collectively sharing information between all companies and positions may introduce unexpected noise and degrade the prediction performance. Prior studies mainly focus on the company- or position-level trend analysis. How to distill commonly shared knowledge and reduce potential noise information for fine-grained company-position demand-supply forecasting is another challenge.
Third, the volume of talent demand and supply timely varies, and forecasting the fine-grained talent demand and supply for multiple companies further strengthens the sparsity issue.
Many companies only have demand and supply records in short periods.
The last challenge is accurately predicting talent demand and supply variation based on a few instances.

To address the aforementioned challenges, in the preliminary paper~\cite{guo2022talent}, we propose the \emph{Dynamic Heterogeneous Graph Enhanced Meta-learning}~(DH-GEM) framework.
Specifically, we first construct fine-grained talent demand-supply sequences and a time-evolving company-position graph to encode the co-evolve patterns of demand-supply sequences and company-position pairs. We devise a \emph{Demand-Supply Joint Encoder-Decoder}~(DSJED) to attentively capture the intrinsic correlation between demand and supply variation.
Moreover, to incorporate the time-evolving relationship between companies and positions, we propose the \emph{Dynamic Company-Position Heterogeneous Graph Convolutional Network}~(DyCP-HGCN) to selectively preserve common knowledge between company and position representations for more effective demand-supply prediction.
Finally, a \emph{Loss-Driven Sampling based Meta-learner}~(LDSM) is proposed to train the prediction framework, in which companies with fewer data are optimized with a higher learning priority to obtain better initial parameters. 
In this way, the long-tail demand-supply prediction tasks can absorb high-level knowledge from companies with sufficient training data to achieve better prediction performance.
DH-GEM has been deployed as a core functional component of the intelligent human resource system of Baidu, providing timely insights and guidance for users.

\new{
However, talent demand and supply data are often siloed in competing companies, making it difficult to access human resource data in the real world.
As human resource data usually contains rich sensitive information, \eg gender, age, employee compensation, benefit, performance evaluations, \etc
Key stakeholders of companies may exhibit reluctance towards disclosing such information out of concern for the potential ramifications, including granting competitors an unfair advantage in the marketplace, exposing companies to potential malevolent attacks, and engendering ethical and legal quandaries concerning privacy and confidentiality~\cite{huang2001effects,hubbard1998human}.

Federated Learning~(FL)~\cite{mcmahan2017communication} has emerged as a promising solution to tackle the issues of data isolation and privacy protection.
In FL, clients hold private data while a trusted server coordinates collaborative model learning without exposing explicit data.
Intuitively, FL is suitable to be applied in inter-company labor market forecasting, where companies participate as clients with their data privacy protected.
In light of this, we formulate the demand-supply joint prediction task as a \textbf{Federated Labor Market Forecasting}~(FedLMF) problem, which aims to construct a precise predictive model based on isolated human resource data and keep the sensitive information inaccessible by any other clients and the server.

Nevertheless, the human resource data distributions of companies diverge. The reasons include divergent company sizes~\cite{delaney1996impact}, geographic locations~\cite{lepak1999human}, human resource strategies~\cite{delery1996modes}, \etc
In consequence, with the deployment of an FL framework, we confront the challenge of the non-independence and non-identical distribution~(Non-IID) issue.
Additionally, the amount of companies vastly surpasses conventional FL applications that feature only a limited number of participants, thereby exacerbating the non-IID situation.

To address the aforementioned challenges, we propose the \textbf{Meta-personalized Convergence-aware Clustered Federated Learning}~(MPCAC-FL) framework to extend DH-GEM to accurately and timely forecasting labor market trends with privacy preservation on decentralized data.
% Specifically, we first propose to learn a set of individual models in a meta-learning paradigm, where each company optimizes a personalized model based on its own data.
Specifically, we improve the LDSM module in DH-GEM under a federated learning setting in two steps. First, clients learn personalized models in a meta-learning paradigm, where the server can obtain initial parameters for all clients by aggregation among locally meta-learned parameters. Second, during the next federated learning round, clients are sampled with a probability proportional to their training loss values.
% Then, we devise the \textbf{Convergence-aware Clustering}~(CAC) to dynamically cluster clients according to current parameters and converging status.
% Independent federated averaging aggregation is applied on each cluster to avoid harm from model heterogeneity for aggregation and stabilize model convergence, thus improving prediction precision.
Then, we devise the \textit{Convergence-aware Clustering}~(CAC) to cluster clients according to model parameters and converging status dynamically. Then, we apply an independent federated averaging aggregation on each cluster to increase model homogeneity in each cluster. In this way, the non-IID issue among heterogeneous clients can be alleviated and the local model can be optimized in a more stable process.
Besides, we regularize local model parameters to encourage clients with different data distributions to learn more generalizable parameters.
Extensive experiments have been conducted to show that MPCAC-FL outperforms federated baseline models on prediction accuracy while maintaining nearly lossless prediction accuracy under privacy restrictions compared to the state-of-the-art non-federated framework.
Compared with the previous work~\cite{guo2022talent}, our contributions are summarized as follows:
\begin{itemize}
    \item We formulate the Federated Labor Market Forecasting problem for company demand and supply joint forecasting. Then we build a Meta-personalized Convergence-aware Clustered Federated Learning framework upon DH-GEM. MPCAC-FL is a collaborative method without explicit human resource data sharing between privacy-aware companies.
    \item On the client side, we devise a meta-learning module for personalized federated learning on heterogeneous companies and apply a local optimizing objective regularization to improve the model generalization.
    \item On the server side, we propose the Convergence-aware Clustering algorithm to adaptively cluster homogeneous clients for more effective federated aggregation, mitigating the negative effects of non-IID issues.
    \item Extensive experiments show that MPCAC-FL outperforms existing federated optimization methods, and achieves comparable performance with state-of-the-art non-federated models.
\end{itemize}
}

\section{Preliminaries}

\subsection{Data Collection and Description}

We collect real-world data from LinkedIn\footnote{LinkedIn Website: \url{https://www.linkedin.com/}}, one of the largest online professional networks (OPNs), where companies can publish job postings for talent hunting and employees can create their own profiles of work experiences.

\begin{figure}[t]
  \centering
  \includegraphics[width=\linewidth]{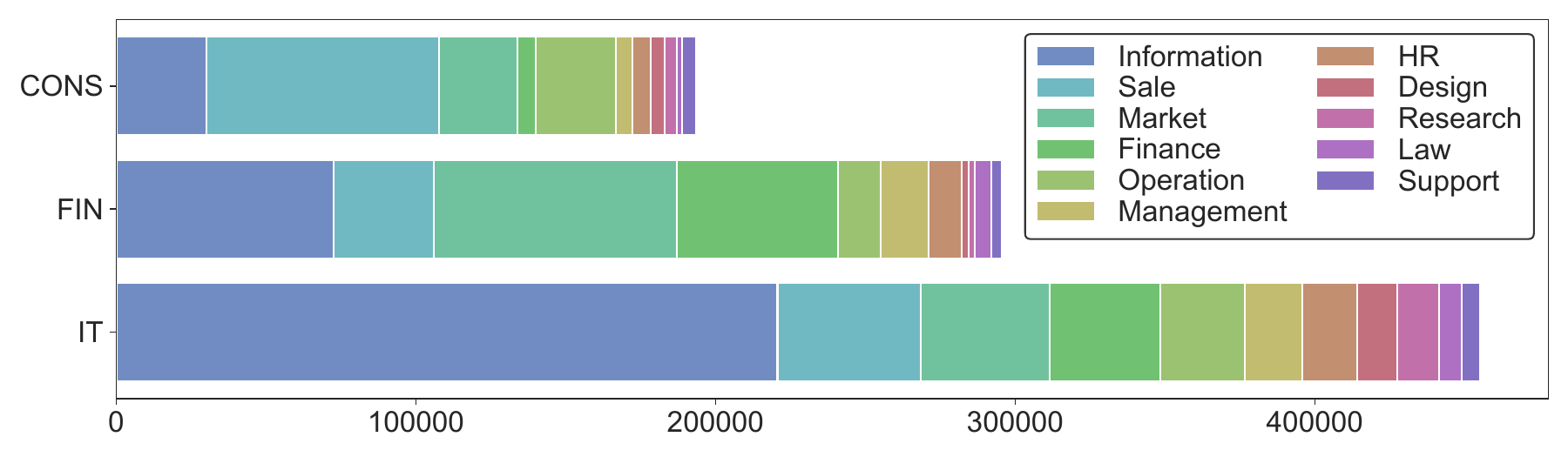}
  \caption{Positions distribution of three real-world datasets.}
  \label{fig:position}
\end{figure}

Specifically, we construct large-scale datasets from three major industries, \ie \emph{Information Technology}~(IT), \emph{Finance}~(FIN), and \emph{Consuming}~(CONS). All three datasets are ranged from March 2016 to March 2019.
Particularly, there are $455,192$ job postings and $2,004,973$ work experiences in IT, $295,651$ job postings and $1,787,386$ work experiences in FIN, and $193,481$ job postings and $1,237,048$ work experiences in CONS.
Following the official position titles for job hunting on LinkedIn and existing techniques~\cite{li2020deep}, we categorize and align raw positions in job posting and work experience data into $11$ classes, including \emph{Information}, \emph{Sale}, \emph{Market}, \emph{Finance}, \emph{Operation}, \emph{Management}, \emph{HR}, \emph{Design}, \emph{Research}, \emph{Law} and \emph{Support}. The distribution of each position is shown in Figure~\ref{fig:position}.

\subsection{Data Preprocessing}
\label{sec:preprocessing}

\new{
For privacy concerns, companies are usually unwilling to disclose their human resource data, which is helpful to preserve competitive advantage, prevent malicious attacks, and avoid privacy issues. Consequently, in this work, we do not access actual human resource data on talent demand and supply. Instead, following previous works~\cite{zhang2021attentive,zhang2021talent} we utilize publicly available job postings and employee work experience data as substitutes.

More precisely, talent demand refers to the number of personnel that a company requires for a specific position at a given point in time, to promote business growth or minimize turnover. When talent demand arises within an organization, job postings are typically released to attract suitable and qualified candidates, leading to a positive correlation between talent demand and corresponding job postings~\cite{yao2013learning}.
On the other hand, talent supply pertains to the number of candidates that a company makes available to the labor market~\cite{makarius2017addressing,aviv2001effect}, primarily in the form of departing employees. We use the job hopping in work experience data to estimate the quantity of talent that a company provides to other organizations.
}
\begin{definition}
    \label{def:demand}
    \textbf{Talent Demand and Talent Demand Sequence.} Talent demand $D_{p}^{t}$ is defined as the number of job postings published by a company for position $p$ at timestamp $t$. Correspondingly, the talent demand sequence is defined as a time-series $D_{p}^{1, T}=\{D_{p}^{t}|1 \leq t \leq T\}$, where $T$ is the length of the sequence.
\end{definition}
\begin{definition}
    \label{def:supply}
    \textbf{Talent Supply and Talent Supply Sequence.} Talent supply $S_{p}^{t}$ is defined as the number of job hopping and position $p$ at timestamp $t$. Correspondingly, the talent supply sequence is defined as a time-series $S_{p}^{1, T}=\{S_{p}^{t}|1 \leq t \leq T\}$, where $T$ is the length of the sequence.
\end{definition}

Building upon earlier research~\cite{zhang2021talent}, we discretize continuous time into a sequence of uniform-length intervals (\ie one month), and align talent demand and supply sequences.
Moreover, we augment the uni-variate sequence by incorporating sequential segmentation and value normalization.
\new{Last, we quantize the prediction values to trend types, for example, we consider five types of trend types, including \emph{sharply increasing}, \emph{steady increasing}, \emph{stable}, \emph{steady decreasing, sharply decreasing}.}
The enhanced talent demand-supply sequences describe the fine-grained labor market trend variation and can be utilized for subsequent talent demand-supply forecasting.

\new{
Furthermore, we use the above datasets for the defined federated problem in Section~\ref{sec:pro}. Specifically, we split the talent demand data according to the publishing companies of job postings and the talent supply data according to the employed companies of work experiences. Each sub-dataset corresponds to one company's talent demand and supply data, which will be held by a unique client in federated learning. Compared with the usual federated learning with only several clients, the large amount of companies intensifies the non-IID issue~\cite{zhu2021federated}. Moreover, we need to create the same number of clients and preserve a large number of optimizing models, which will bring training burdens. Due to limited computational resources, we restrict our selection to the top $100$ data-sufficient companies for the IT, FIN, and CONS datasets, respectively.
}

\begin{figure*}[t]
    \centering
    \subfigure[Overall distribution over time.] {\includegraphics[width=0.24\textwidth]{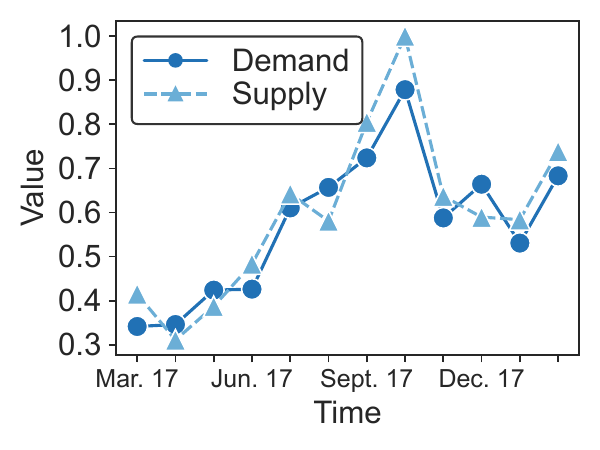}\label{fig:overall_trend}}
    \subfigure[Joint talent demand-supply distribution.] {\includegraphics[width=0.24\textwidth]{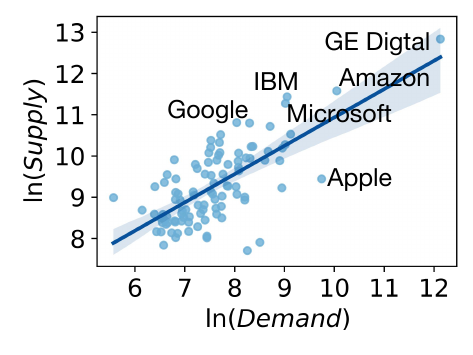}\label{fig:company_dist}}
    \subfigure[Correlation of connected companies in DyCP-HG.] {\includegraphics[width=.24\linewidth]{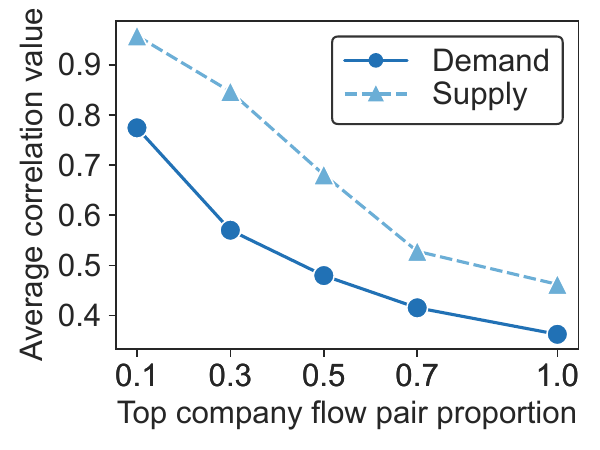} \label{fig:graph_corr}}
    \subfigure[Normalized volume distribution.] {\includegraphics[width=.24\linewidth]{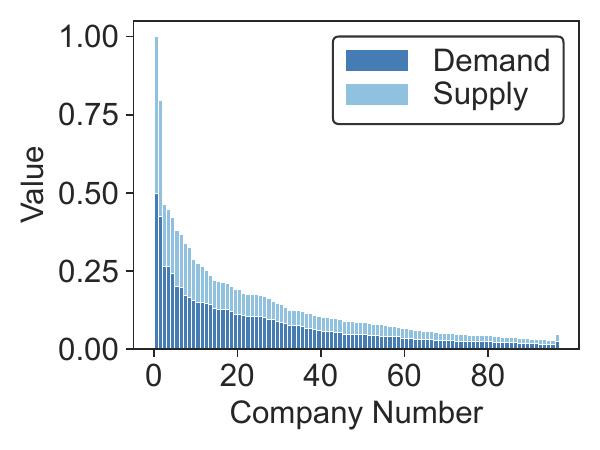} \label{fig:demand_supply_dist}}
    \vspace{-3pt}
    \caption{Distributions of the IT dataset:
        (a) the overall talent demand and supply distribution,
        (b) joint talent demand-supply distribution of companies,
        (c) normalized volume distribution of talent demand and supply,
        (d) correlation of talent demand and supply between connected companies and positions in DyCP-HG.
    }
  \label{fig:it}
\end{figure*}

\subsection{Data Exploration}

This section presents an analysis of 1)~the correlation between demand and supply sequences, 2)~the relationship between companies and positions, and 3)~non-IID distribution on the IT dataset. Similar distributions were observed for other datasets, but due to space constraints, we do not report them here.

\textbf{Correlation between talent demand and supply trends.}
\label{sec:demandsupplyquan}
We conduct a preliminary analysis of the talent demand and supply sequences. Our analysis, depicted in Figures~\ref{fig:overall_trend} and~\ref{fig:company_dist}, indicates that talent demand and supply are positively correlated and display significant variation over time. These findings motivate us to explore joint prediction strategies.

\new{\textbf{Relationship between companies and positions.}
Figure~\ref{fig:graph_corr} exhibits the mean Pearson correlation among companies. We note a substantial rise in the correlation of talent demand and supply sequences as we adjust $k$, the proportion of company pairs with the highest job hopping rate between them, from $100\%$ to $10\%$. This indicates that the correlation between companies can offer valuable insights for predicting talent demand and supply trends.

\textbf{Non-independence and non-identical distribution of talent demand and supply values.}
\label{sec:analysis_longtail}
Figure~\ref{fig:demand_supply_dist} presents the normalized volume distribution of talent demand and supply. It reveals a highly synchronized long-tail distribution of talent demand and supply, where the demand-supply volume of over $80\%$ companies is less than $0.25$. This distribution leads to a non-IID challenge due to different companies' diverse industrial maturity and employment scope.}

\begin{figure*}
    \centering
    \includegraphics[width=\textwidth]{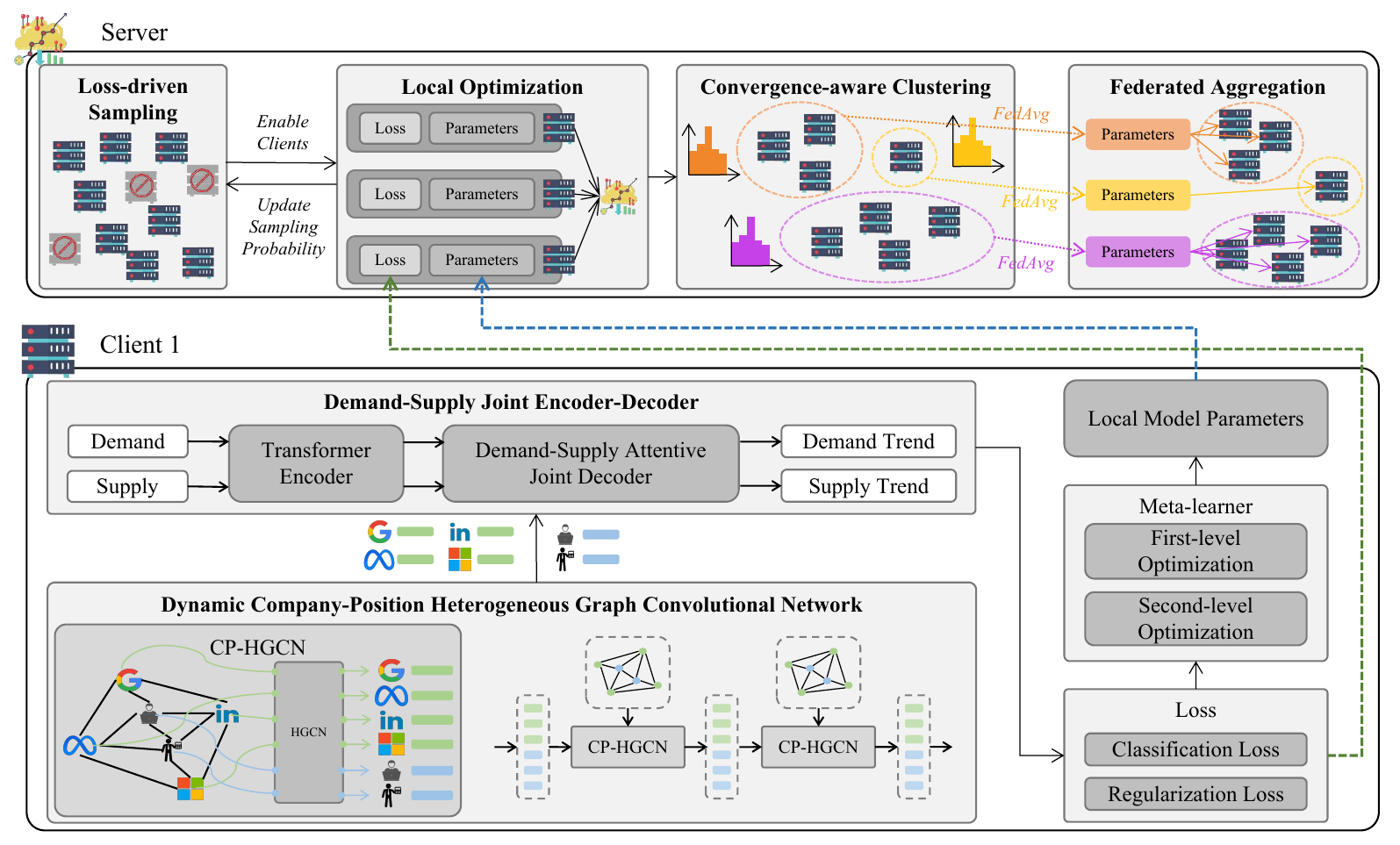}
    \caption{\new{An overview of Meta-personalized Convergence-aware Clustered Federated Learning framework.}}
    \label{fig:mpcacfl}
\end{figure*}

\new{

\section{Framework}

\subsection{Problem Definition}\label{sec:pro}
To learn the labor market forecasting models for multiple companies, we establish a collaborative learning architecture, where a trusted server coordinates clients that are acted by companies. We propose to study the Federated Labor Market Forecasting (FedLMF) problem, which aims to optimize the models through inter-company federated learning without explicitly sharing data. Specifically, we define FedLMF in Problem~\ref{pro:1}.

\begin{problem}
    \label{pro:1}
    \textbf{Federated Labor Market Forecasting.} For client $i$, given the talent demand and supply sequences $D^{1, T}_{c, p}$ and $S^{1, T}_{c, p}$ of the company-position pair $(c, p)$, we aim to simultaneously predict demand and supply for pair $(c, p)$ in the next timestamp, as
    \begin{equation}
    y_D^{T+1}, y_S^{T+1} \gets \mathcal{F}_{\theta_i}{(D^{1, T}_{c, p}; S^{1, T}_{c, p}}),
    \end{equation}
    where $y_D^{T+1}$ and $y_S^{T+1}$ are the estimated company-position-wise talent demand and supply trend in the next timestamp, and $\mathcal{F}_{\theta_i}(\cdot)$ is the parameterized joint prediction function. We aim to learn $\mathcal{F}_{\theta_i}(\cdot)$ for each client $i$ to achieve the best predictive accuracy on average among all clients.
\end{problem}

By default, the model structure and the training scheme remain the same across all clients.

\subsection{Overview}

Figure~\ref{fig:mpcacfl} illustrates the Meta-personalized Convergence-aware Clustered Federated Learning (MPCAC-FL) framework, where companies participate in federated learning as clients. Under this centralized architecture, clients preserve their local models and private data, and the server has a federated optimization algorithm that coordinates the training process with clients.

We first introduce the basic prediction model, Dynamic Heterogeneous Graph Enhanced Meta-learning~(DH-GEM), with two modules (\ie Demand-Supply Joint Encoder-Decoder~(DSJED) in Section~\ref{sec:dsjed} and Dynamic Company-Position Heterogeneous Graph Convolutional Network~(DyCP-HGCN) in Section~\ref{sec:graph}).
Then the loss function and model optimization scheme of DH-GEM are specified in Section~\ref{sec:loss} and Section~\ref{sec:optimize}, respectively.
Moreover, we propose the Convergence-aware Clustered Federated Learning~(CAC-FL) algorithm executed in the server for effective federated optimization in Section~\ref{sec:cacfl}.
Last, we generally present the full training pipeline in Section~\ref{sec:train} to illustrate how clients optimize models with the coordination of a server.

}

\subsection{Dynamic Heterogeneous Graph Enhanced Meta-learning}

This subsection introduces the client-side modules, including \emph{Demand-Supply Joint Encoder-Decoder}~(DSJED), \emph{Dynamic Company-Position Heterogeneous Graph Convolutional Network}~(DyCP-HGCN), loss functions for classification and generalization, and meta-learning for personalization.

\subsubsection{Demand-supply joint encoder-decoder.}
\label{sec:dsjed}

The intrinsic correlation between talent demand and supply is discovered in our analysis presented in Section~\ref{sec:demandsupplyquan}. In this light, we design our novel \emph{Demand-Supply Joint Encoder-Decoder}~(DSJED) to seek to highlight the inherent correlation between demand and supply, which delivers enhanced predictions for both demand and supply with its attentive mechanism.

To adequately represent each element of the demand and supply sequences for sequential modeling, we utilize a multi-layer perceptron to map each scalar value to a high-dimensional vector. To further enhance the fidelity of our model, we incorporate information related to the specific company and position by concatenating these vectors with their corresponding temporal company and position embeddings, denoted as $h_c^{t}$ and $h_p^{t}$. Subsequently, we pass the concatenated vectors through another multi-layer perceptron to obtain the optimal representation of each element in the demand and supply sequences, \ie $h_D^{t}$ and $h_S^{t}$. Section~\ref{sec:graph} will elaborate on the construction of the aforementioned temporal company and position embeddings.

Subsequently, in order to obtain a comprehensive representation of both the demand and supply sequences, we utilize the encoder of Transformer~\cite{vaswani2017attention} equipped with sinusoidal positional encoding. Notably, the parameters of the encoder are shared between the two sequences. This approach of using an identical mapping for both sequences proves to be advantageous in capturing any common evolving patterns which may exist in both demand and supply. In such a way, the trend embedding of demand $h_D$ and supply $h_S$ can be effectively represented through the fusion of $\{h^{1}, \cdots, h^{T}\}$ using a Transformer encoder.

\begin{figure}[t]
    \centering
    \includegraphics[width=\linewidth]{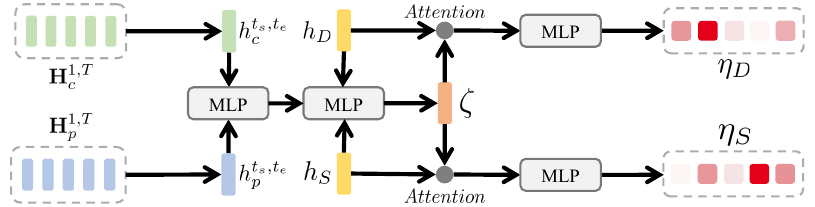}
    \caption{The architecture of DSAJD.}
    \label{fig:dsajd}
\end{figure}

Based on the strong correlation between demand and supply, which was discussed in Section~\ref{sec:demandsupplyquan}, we propose the \emph{Demand-Supply Attentive Joint Decoder}~(DSAJD) shown in Figure~\ref{fig:dsajd}, with the aim of effectively decoding the sequential encodings for both demand and supply, while taking into account their mutual relationships. \new{Initially, we generate the time-evolving embedding between the global timestamp\footnote{\new{We denote timestamps as $1,\cdots,T$ for any particular demand or supply sequence during training, and timestamps of different sequences correspond to different global timestamps due to the sequence sampling in data argumentation introduced in Section~\ref{sec:preprocessing}}.} $t_1$ and $t_2$ for company $c$ and position $p$, respectively, as follows
\begin{equation}
    \begin{array}{c}
        h^{t_1, t_2} = \mathbf{M}^{t_1, t_2} \cdot \mathbf{A} \cdot \mathbf{H},
    \end{array}
\end{equation}
where $h^{t_1, t_2}$ can be $h_c^{t_1, t_2}$ or $h_p{t_1, t_2}$, $\mathbf{M}_{t_1, t_2}$ is a $0$-$1$ vector that indicates whose indices between $t_1$ and $t_2$ as one, $\mathbf{A}$ is a learnable attentive vector and $\mathbf{H}$ is the list of $(\cdots,h_{c}^{t_1},\cdots, h_{c}^{t_2},\cdots)$ or $(\cdots,h_{p}^{t_1},\cdots, h_{p}^{t_2},\cdots)$.}
To create the company-position-aware demand-supply joint sequential feature $\zeta$, we fuse $h_D$, $h_S$, $h_c^{1, T}$ and $h_p^{1, T}$ through
\begin{equation}
    \zeta = \operatorname{MLP}(h_D||h_S||\operatorname{MLP}(h_c^{1, T}||h_p^{1, T})),
\end{equation}
where $\operatorname{MLP}(\cdot)$ represents the multi-layer perceptron, the $\cdot||\cdot$ represents the concatenation operation.
In addition, in order to facilitate effective information sharing, we incorporate two attentive modules that merge the features of $\zeta$ with $h_D$ and $h_S$, respectively, resulting in the formation of new features denoted as $\hat{h_D}$ and $\hat{h_S}$, expressed as $\hat{h} = \mathbf{w}[h; \zeta]$, where $h$ can be $h_D$ or $h_S$, $\hat{h}$ can be $\hat{h_D}$ or $\hat{h_S}$, $\mathbf{w}$ denotes learnable parameters.
Finally, $\hat{h_D}$ and $\hat{h_S}$ are separately fed into two independent multi-layer perceptrons. The output is then operated by $\operatorname{LogSoftMax}$ as vector $\eta_D$ and $\eta_S$. Specifically, the output vector dimension is equal to the number of trend types, and the $i$-th element is the predicted probability of the trend type $i$. Moreover, we use $\operatorname{ArgMax}$ to transform $\eta_D$ and $\eta_S$ into the trend type $y_D$ and $y_S$ as follows $y = {\arg\max}_{i\in [1,|Y|]} \eta^i$, where $y$ is $y_D$ or $y_S$, $\eta$ is $\eta_D$ or $\eta_S$, $\eta^i$ is the $i$-th element of $\eta$, and $|Y|$ denotes the number of trend types.

\subsubsection{Dynamic company-position representation learning.}
\label{sec:graph}

\new{
To enhance the prediction accuracy of labor market trends, we aim to learn company and position representations across varied timestamps by incorporating the dynamic relationship between companies and positions.
}
We first construct the Dynamic Company-Position Heterogeneous Graph~(DyCP-HG) by extracting the job-hopping information from the work experiences data, which serves as a powerful module for capturing the co-evolving patterns and intricate interrelationships between companies and positions over time.

\begin{definition}
    \label{def:cphg}
    \textbf{Company-Position Heterogeneous Graph}~(CP-HG). The Company-Position Heterogeneous Graph is defined as $G = (V, E)$, where $V = V_C \cap V_P$ and $E = E_{c, c} \cap E_{p, p} \cap E_{c, p}$. $V_C$ and $V_P$ are nodes of all companies and positions.
\end{definition}

\begin{definition}
    \textbf{Dynamic Company-Position Heterogeneous Graph}~(DyCP-HG). The Dynamic Company-Position Heterogeneous Graph is defined as $\mathcal{G}^{1, T}=(G^{1}, \cdots, G^{T})$ where $1$ and $T$ are the start and end timestamp, and $G^t$ is a CP-HG at timestamp $t$ satisfying $1 \leq t \leq T$.
\end{definition}

On the one hand, the frequent job-hopping between companies and positions describes a high relevance between company- and position-pairs, which can positively influence the labor market trend~\cite{boschma2014labour}. 
On the other hand, the preserved heterogeneous relationship between companies and positions described by the edge connection also provides extra information for prediction~\cite{guerrero2013employment}.

\begin{figure}[t]
    \centering
    \includegraphics[width=\linewidth]{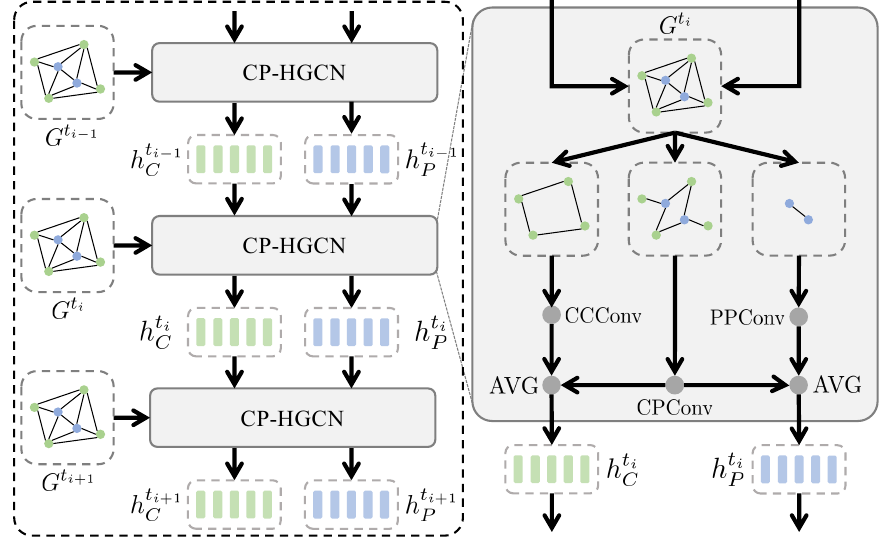}
    \caption{The architecture of DyCP-HGCN.}
    \label{fig:dycphgcn}
\end{figure}

To transform the knowledge involved in the dynamic company-position heterogeneous graphs, we propose the \emph{Dynamic Company-Position Heterogeneous Graph Convolutional Network}~(DyCP-HGCN) to optimize companies and positions representations on it. As shown in Figure~\ref{fig:dycphgcn}, the DyCP-HGCN leverages a dynamic recurrent process to encode the DyCP-HG with the output of node embedding for each timestamp.
We denote DyCP-HGCN as $\Phi$ and define it as
\begin{equation}
    (\mathbf{H}_C^{1}, \cdots, \mathbf{H}_C^{T}), (\mathbf{H}_P^{1}, \cdots, \mathbf{H}_P^{T}) = \Phi(\mathcal{G}^{1, T}),
\end{equation}
where $\mathbf{H}_C^{t}$ and $\mathbf{H}_P^{t}$ are the company and position node embedding at timestamp $t$.

\emph{Company-Position Heterogeneous Graph Convolutional Network}~(CP-HGCN) is designed as a cell of DyCP-HGCN to learn the static company and position embedding $h_c$ and $h_p$. We generally define CP-HGCN $\phi(\cdot)$ as follows
% and generally defined as
\begin{equation}
    \mathbf{H}_C, \mathbf{H}_P = \phi(G; \mathbf{H}_C^{\prime}; \mathbf{H}_P^{\prime}),
\end{equation}
where $\mathbf{H}_C^{\prime}$ and $\mathbf{H}_P^{\prime}$ are input company and position node embedding, $\mathbf{H}_C$ and $\mathbf{H}_P$ are output ones. Specifically, $\phi(\cdot)$ contains three steps. 

Firstly, to handle the heterogeneity, we separate a CP-HG $G$ into three sub-graphs according to edge types, \ie $G(V_c, E_{c, c})$, $G(V_p, E_{p, p})$ and $G(V, E_{c, p})$ respectively.

Secondly, for three sub-graphs of demand-supply edges, \ie $E_{c,p}$, company-hopping edges $E_{c,c}$, and position-hopping edges $E_{p,p}$, we adopt three graph convolutional operations, \ie $\operatorname{CPConv}(\cdot)$, $\operatorname{CCConv}(\cdot)$ and $\operatorname{PPConv}(\cdot)$, respectively, to generate the node representation by aggregating the neighboring information. Three convolutional operations can be uniformly defined as
\begin{equation}
    h_{u} =\sigma(\mathbf{b} + \sum\limits_{v \in \mathcal{N}_u}\frac{w_{uv}}{\sqrt{|\mathcal{N}_u|}\sqrt{|\mathcal{N}_v|}}\cdot(h_v \cdot \mathbf{W})),
\end{equation}
where we denote $h_{u}$ as the embedding of node $u$, $\sigma$ as $\operatorname{ReLU}$ activation function, $\mathbf{b}$ and $\mathbf{W}$ as learnable parameters, $w_{uv}$ as the edge weight between node $u$ and $v$, and $|\mathcal{N}_v|$ as the number of neighbors of node $v$.

Thirdly, considering the company node embedding $\mathbf{H}_C$ is produced by $\operatorname{CPConv}(\cdot)$ and $\operatorname{CCConv}(\cdot)$, while  position node embedding $\mathbf{H}_P$ by $\operatorname{CPConv}(\cdot)$ and $\operatorname{PPConv}(\cdot)$, we leverage the meaning operation for company and position embedding respectively to obtain the final output embedding of $\phi(\cdot)$.

Besides, we adopt the recurrent process as shown in Figure~\ref{fig:dycphgcn} to continuously learn the temporal pattern of company and position embedding and generate representations for company and position at each timestamp. Specifically, we use the learned embedding of the previous timestamp $t-1$ as the input of a recurrent cell (\ie CP-HGCN $\phi(\cdot)$) and output the new embedding at the current timestamp $t$. Based on the single cell function $\phi(\cdot)$, the recurrent process at timestamp $t$ is
\begin{equation}
    \mathbf{H}_C^t, \mathbf{H}_P^t = \phi(G^{t}; \mathbf{H}_C^{t-1}; \mathbf{H}_P^{t-1}), 1 \leq t \leq T.
\end{equation}
The $\mathbf{H}_C^{0}$ and $\mathbf{H}_P^{0}$ are initialized randomly. In this way, we can get the list of company and position embedding from $1$ to $T$ orderly, \ie $(\mathbf{H}_C^{1}, \cdots, \mathbf{H}_C^{T}), (\mathbf{H}_P^{1}, \cdots, \mathbf{H}_P^{T})$.

\new{
\subsubsection{Loss Functions}
\label{sec:loss}

To learn a predictive model to classify the labor market trend types, we leverage classification loss $\mathcal{L}^c$ to optimize our model and regularization loss $\mathcal{L}^r$ to preserve generalization ability under personalized requirements. To this end, we define the loss function as $\mathcal{L} = \mathcal{L}^c + \lambda\mathcal{L}^r$. We will give a specific formulation of two loss functions.
}

\textbf{Classification loss function.}
We optimize the model by Negative Log-Likelihood Loss with Poisson distribution (\ie $P(Y=y)=\frac{\eta^{y}}{{y}!}\exp{(-\eta)}$) as
\begin{equation}\label{eq:loss}
    \mathcal{L}^c=\sum{-\log{P(Y=y)}}=\sum \exp{(-\eta)}-y\eta+\log{y!},
\end{equation}
where $y$ represents $y_D$ or $y_S$ and $\eta$ represents $\eta_D$ or $\eta_S$. To simplify the calculation, the last term can be approximated according to the \emph{Stirling's Formula}, $\log{y!} \approx y\log{y}-y+\frac{1}{2}\log({2\pi y})$. 
The overall classification loss is the combination of both the demand and supply prediction loss $\mathcal{L}^{c} = \mathcal{L}^c_{D}+\mathcal{L}^c_S$.

\new{\textbf{Regularization loss function.}
Each client optimizes parameters independently, where the knowledge absorbed into models is represented in different high-dimensional spaces. However, it affects the generalization ability between large-scale clients if all parameters are learned without any prior assumption. To improve the generalization on heterogeneous clients, inspired by FedProx~\cite{li2020federated}, we leverage regularization on partial local models and optimized independently as
\begin{equation}
  \mathcal{L}^r = \frac{\mu}{2}\Vert \tilde{\theta}_{k} - \tilde{\theta}_{k-1} \Vert^2,
\end{equation}
where $\mu$ is a hyperparameter and $\tilde{\theta}$ is the partial parameters which we use the average of dynamic position embeddings $\mathbf{H}_P$ to represent, and $k$ is the current local optimization epoch.}

\new{
\subsubsection{Model-agnostic Meta-learning for Personalized Optimization}
\label{sec:optimize}

The optimization objective is to minimize the loss values, whose function can be defined as
$L(\{\theta_i\}) = \frac{1}{N}\sum^N_{i=1}L_i(\theta_i)$,
where $\{\theta_i\}$ denotes all local model parameters, $N$ denotes the number of active client, and
\begin{equation}
  L_i(\theta_i) =  \mathcal{L}(\theta_i-\alpha\nabla \mathcal{L}(\theta_i)),
\end{equation}
where $\alpha$ is the learning rate of gradient descent.
However, the amount of human resource data in most companies is limited. For models training on limited data, optimal learning directions are difficult to estimate accurately, which is harmful to convergence.
Inspired by the recent success of Model-Agnostic Meta-Learning~(MAML)~\cite{finn2017model} on learning suitable parameter initialization for personalized performance, we extract globally shared meta-knowledge from diverse companies to enable fast adaptation and more accurate predictions when forecasting demand-supply for companies with limited data.
During execution, we first compute the gradients of $L_i(\theta_i)$ as
\begin{equation}
  \nabla L_i(\theta_i) = (I-\alpha\nabla^2\mathcal{L}(\theta_i))\nabla \mathcal{L}(\theta_i-\alpha\nabla \mathcal{L}(\theta_i)),
\end{equation}
where $\alpha$ is the learning rate.
Then we leverage bi-level optimization to update $\theta_i$ as
\begin{equation}
  \begin{array}{rl}
    \theta_i^{(n+1)} &= \theta_i^{(n)}-\beta\nabla L_i(\theta_i^{(n)}),\\
    &= {\theta_i^{(n)}} - \beta(I-\alpha\nabla^2\mathcal{L}(\theta_i^{(n)}))\nabla \mathcal{L}(\theta_i-\alpha\nabla \mathcal{L}(\theta_i)),
  \end{array}
\end{equation}
where $\beta$ is the meta-learning rate for the second-level optimization.
In this way, we can improve the optimization effectiveness for longtail companies with few-shot human resource data, which helps the overall prediction in the market.
}

\new{

\subsection{Convergence-aware Clustered Federated Learning}
\label{sec:cacfl}

As discussed in Section~\ref{sec:analysis_longtail}, companies' human resource data distributions are divergent, causing the non-IID issue to impede the improvement of federated learning performance.
Therefore, we propose the Convergence-aware Clustered Federated Learning~(CAC-FL) framework to address this challenge.
Convergence-aware Clustering~(CAC) is the core component of CAC-FL, which forms groups of homogeneous clients, \ie holding similar local model parameters, by column-pivoted QR factorization~\cite{damle2019simple} enhanced spectral clustering. And we apply FedAvg~\cite{mcmahan2017communication} on each group to aggregate parameters.
Grouping homogeneous clients can facilitate more effective federated aggregation and eliminate harmful information exchange from heterogeneous ones, because aggregating updates from heterogeneous models, \ie holding dissimilar local model parameters, does not benefit them in optimizing toward the optimal direction.

We then specifically introduce the CAC-FL in Algorithm~\ref{alg:cac}, which outputs updated parameters based on the latest parameters, the number of epochs, and loss values.
CAC-FL has four steps, including determining cluster cohesion, clustering clients, federated optimization, and sampling probability update. We will illustrate these steps specifically as follows.

\begin{algorithm}[t]
  \new{
  \caption{Convergence-aware Clustering.}
  \label{alg:cac}
  \LinesNumbered
  % \KwIn{Local parameters $\{\theta^{(n)}_i\}$, the number of training round $n$, and loss values $\{\mathcal{L}_i\}$.}
  % \KwOut{Update local parameters $\{\theta^{(n+1)}_i\}$.}
  \Fn(){\CACFL{$\{\theta^{(n)}_i\}, n, \{\mathcal{L}_i\}$}}{
    \tcc{Decide the number of clusters}
    Determine the loss value distribution $\mathcal{N}(\mu, \sigma^2)$ of all clients in past $\tau$ epochs as Equation~\ref{equ:dist}\;
    Estimate the convergence degree $\varrho$ as Equation~\ref{equ:cvgdegree}\;
    Decide cluster number $m$ based on $\varrho$ as Equation~\ref{equ:alpha}\;
    \tcc{Spectral clustering on clients}
    Compute affinity matrix $\mathbf{S}$ based on client features $\{\tilde{\theta}^{(n)}_i\}$ as Equation~\ref{equ:affinity}\;
    Compute normalized Laplacian Matrix $\mathbf{L}$ based on adjacency matrix $\mathbf{W}$ and degree matrix $\mathbf{D}$ derived from affinity matrix $\mathbf{S}$\;
    Compute the first $k$ eigenvectors $v_1, \cdots, v_{k}$ of $\mathbf{L}$\;
    Form the matrix $\mathbf{V}$ by stacking eigenvectors $v_1, \cdots, v_{k}$ on top of each other\;
    Cluster the rows of $\mathbf{V}$ using column-pivoted QR factorization into $m$ clusters\;
    \tcc{Use FedAvg on each cluster}
    \ForEach{cluster of clients $\tilde{\mathbb{C}}$}{
      Aggregate parameters $\theta^{(n+1)}_{\tilde{\mathbb{C}}} \gets \frac{1}{N}\sum^{i=1}_{N}{\theta^{(n+1)}_i}$\;
      Update parameters $\theta^{(n+1)}_i \gets \theta^{(n+1)}_{\tilde{\mathbb{C}}}$\;
    }
    \Return{\text{Update local parameters }$\{\theta^{(n+1)}_i\}$}
  }
  }
\end{algorithm}

\textbf{1)~Convergence-aware adaptive cluster cohesion determination.}
In the first step, we aim to decide the number of clusters $m$ first, which affects the homogeneity of clients within each cluster.
Homogeneity among a group of clients is beneficial for federated aggregation to avoid client shifts from heterogeneous ones. Meanwhile, intuitively, clients tend to be more homogeneous in a cluster if their amount is smaller. Therefore, forming smaller clusters can increase homogeneity to help federated aggregation.
Moreover, when the training approaches convergence, local models are optimized toward certain directions and sensitive to large parameter transformation during federated aggregation. This fact motivates us to form fewer clusters for stabler convergence when the optimization comes to an end.
Therefore, we dynamically decide $m$ by measuring the convergence degree $\varrho$ using the preceding optimization results.
We usually regard that training converges when errors no longer exhibit a significant decrease but fluctuate within a small range.
To describe this small range in the past $\tau$ rounds, we calculate the normal distribution of global training loss values $\mathcal{N}(\mu, \sigma^2)$ as
\begin{equation}
\label{equ:dist}
    \mu = \frac{1}{\tau}\sum\limits^{j=1}_{\tau}\mathcal{L}^{(n-j)},\ \ \sigma^2 = \frac{1}{\tau}\sum\limits^{j=1}_{\tau}(\mathcal{L}^{(n-j)} - \mu)^2,
\end{equation}
where $n$ is the training epoch index.
Then, to evaluate the current fluctuation of loss values, we compute the appearance probability of $\mathcal{L}^{(n)}$ in $\mathcal{N}(\mu, \sigma^2)$ as the convergence degree $\varrho$ by
\begin{equation}
  \label{equ:cvgdegree}
  \varrho = 1 - 2|\int_{\mu}^{\mathcal{L}^{(n)}}\mathcal{N}(\mu, \sigma^2)dx|,
\end{equation}
where the lower value of $\varrho$ means that the loss stops decreasing and fluctuating, \ie the model converges.
In this way, $\varrho$ can further support the calculation of $m$ as
\begin{equation}
  \label{equ:alpha}
  m = \min(1 + \sqrt{n}e^{\varrho}, |\mathbb{C}|),
\end{equation}
where $m$ is limited below the client number $|\mathbb{C}|$.
To ensure the reliability of the decision of $\alpha$, we activate this convergence measurement after a manually set number of rounds.

\begin{algorithm}[t]
  \new{
  \caption{Meta-personalized Convergence-aware Clustered Federated Learning.}
  \label{alg:mpcacfl}
  \LinesNumbered
  \KwIn{client set $\mathbb{C} = \{C_i|1 \leq i \leq |\mathbb{C}|\}$, model $\mathcal{F}(\cdot)$, and sampling rate $r$.}
  \KwOut{model parameters $\theta_i$ for each client $C_i \in \mathbb{C}$.}
  \Fn(){\MPCACFL{$\mathbb{C}, \mathcal{F}$, $r$}}{
    Initialize model parameters $\theta^{(0)}$\;
    \While{not converged}{
      Sample $N = |\mathbb{C}| * r$ clients as $\hat{\mathbb{C}}$ according the sampling probability $\{p_i\}$ of client $i$\;
      \For{$C_i \in \hat{\mathbb{C}}$ in parallel}{
        Compute $\mathcal{L}^{(n)}_i$ based on given local dataset $(X_i, Y_i)$ by two loss functions\;
        Calculate model gradients $\nabla\mathcal{L}^{(n)}_i$\;
        Optimize $\theta^{(n)}_i$ with $\nabla{\mathcal{L}^{(n)}_i}$ by Meta-personalization\;
      }
      Update all local parameters $\{\theta^{(n+1)}_i\}$ by \CACFL{$\{\theta_i\}, n, \{\mathcal{L}^{(n)}_i\}$}\;
      Update sampling probability $\{p_i\}$ based on $\{\mathcal{L}_i\}$
    }
    \Return{$\{\theta_i\}$}
  }
  }
\end{algorithm}

\textbf{2)~CPQR factorization-based spectral clustering among heterogeneous models.}
Secondly, we cluster clients to aggregate based on pairwise similarity for forming homogeneous clients.
We feature client $C_i$ by $\tilde{\theta}^{(n)}_i$ through collecting position embedding weights from local models. To enable spectral clustering~\cite{ng2001spectral} among clients, we first compute the affinity matrix $\mathbf{S} = \{s_{i, j}\}$ via the radian basis function~(RBF) as
\begin{equation}
  \label{equ:affinity}
  s_{i, j} = \exp(-\frac{\Vert \tilde{\theta}^{(n)}_i-\tilde{\theta}^{(n)}_j \Vert_2^2}{2\sigma^2}),
\end{equation}
where $\sigma$ is a hyper-parameter.
Then, we compute the Laplacian Matrix $\mathbf{L} = \mathbf{D}^{-\frac{1}{2}}(\mathbf{D}-\mathbf{W})\mathbf{D}^{-\frac{1}{2}}$.
The first $k$ decomposed eigenvectors $v_1, \cdots, v_{k}$ of the matrix $\mathbf{L}$ will be used for forming the matrix $\mathbf{V}$ by stacking eigenvectors on top of each other.
We follow an efficient method for spectral clustering based on column-pivoted QR~(CPQR) factorization~\cite{damle2019simple} to decompose the matrix $\mathbf{V} \in \mathbb{R}^{k \times N}$ as
\begin{equation}
  \label{equ:cpqr1}
  \mathbf{V}\mathbf{P} = \mathbf{Q}\mathbf{R},
\end{equation}
where $\mathbf{Q} \in \mathbb{R}^{k \times k}$ is unitary, $\mathbf{R} \in \mathbb{R}^{k \times N}$ is upper triangular, and $\mathbf{P} \in \mathbb{R}^{N \times N}$ is a column permutation matrix.
We denote $\mathbf{P}_m$ as the first $m$ columns of $\mathbf{P}$, which is then transformed through the polar factorization~\cite{brenier1991polar} as
\begin{equation}
  \label{equ:cpqr2}
  \mathbf{P}_m = \mathbf{U}\mathbf{H},
\end{equation}
where $\mathbf{U} \in \mathbb{R}^{m \times m}$ is orthogonal and $\mathbf{H} \in \mathbb{R}^{m \times m}$ is positive semi-definite.
Client $i \in [1, N]$ assigned with its corresponding cluster label $\operatorname{Cluster}_i = \arg\max_j(|\mathbf{U}^T\mathbf{V}|_{j,i})$ .

\textbf{3)~Federated optimization across homogeneous clients.}
Next, to federally optimize models in homogeneous clients, we apply FedAvg on each cluster to aggregate and update local model parameters as
\begin{equation}
\theta^{(n+1)}_{\tilde{\mathbb{C}}} \gets \frac{1}{|\tilde{\mathbb{C}}|}\sum^{i=1}_{|\tilde{\mathbb{C}}|}{\theta^{(n+1)}_i}.
\end{equation}

\textbf{4)~Loss-driven sampling probability update.}
Last, intuitively, the client with higher training loss indicates a larger prediction error, thus requiring additional learning efforts. Therefore, after clients update their parameters, we update the sampling probability of clients by
\begin{equation}
    p_i^{(n+1)}=\frac{e^{\mathcal{L}_i^{n}}}{\sum_{j=1}^{N} e^{\mathcal{L}_j^{n}}}
\end{equation}
where $p_c^{(n+1)}$ is the sampling probability for round $n+1$, $\mathcal{L}_j^{n}$ is the validated loss in round $n$.

\subsection{Federated Training Workflow}\label{sec:train}
The federated training workflow is reported in Algorithm~\ref{alg:mpcacfl}.
In each round of federated optimization, the server samples active clients by loss-driven sampling probability.
After that, each client computes gradients on local data and optimizes with meta-learning-based personalization.
Then the server recursively updates the local parameters collected from active clients by convergence-aware clustered federated learning.
The sampling probability will be updated based on loss values for the sampling of the next round.
This iterative process continues until the training converges, and the parameters for each local model will be returned.
}

\begin{table*}[t]
  \centering
  \caption{\new{The overall performance of labor market forecasting on three real-world datasets for baselines optimized via \textsc{Global}, \textsc{Local}, and \textsc{Federated}. 
  \underline{Underlined}: the upper bound of the state-of-the-art model without federated settings.
  \textbf{Bold}: the best performance below the upper bound.}}
  \label{tab:overall}
  \new{\begin{tabular}{c|ccc|ccc|ccc}
      \toprule
      \multirow{2}{*}{Model} & \multicolumn{3}{c|}{IT} & \multicolumn{3}{c|}{FIN} & \multicolumn{3}{c}{CONS}\\
      & Accuracy & Weighted-F1 & AUROC & Accuracy & Weighted-F1 & AUROC & Accuracy & Weighted-F1 & AUROC\\
      \midrule
      \multicolumn{10}{c}{\textsc{Global}}\\
      \midrule
      LV & 0.3750 & 0.3019 & 0.6690 & 0.3906 & 0.3331 & 0.6831 & 0.3739 & 0.2999 & 0.6767 \\
      LR & 0.5099 & 0.4776 & 0.8011 & 0.5250 & 0.5078 & 0.8100 & 0.4962 & 0.4812 & 0.7899 \\
      GBDT & 0.6134 & 0.6083 & 0.8778 & 0.5981 & 0.5938 & 0.8683 & 0.5469 & 0.5398 & 0.8344 \\
      LSTM & 0.6034 & 0.5995 & 0.8732 & 0.6001 & 0.5860 & 0.8697 & 0.5632 & 0.5581 & 0.8458 \\
      Transformer & 0.6343 & 0.6375 & 0.8950 & 0.6191 & 0.6180 & 0.8842 & 0.5737 & 0.5726 & 0.8551 \\
      \underline{DH-GEM} & \underline{0.6813} & \underline{0.6840} & \underline{0.9168} & \underline{0.6791} & \underline{0.6825} & \underline{0.9155} & \underline{0.6230} & \underline{0.6249} & \underline{0.8883} \\
      \midrule
      \multicolumn{10}{c}{\textsc{Local}}\\
      \midrule
      Local-LV & 0.3042 & 0.2948 & 0.5932 & 0.3490 & 0.2839 & 0.6193 & 0.3294 & 0.2534 & 0.6012 \\
      Local-LR & 0.4738 & 0.3984 & 0.7023 & 0.4293 & 0.4634 & 0.7403 & 0.4893 & 0.4734 & 0.7534 \\
      Local-GBDT & 0.5023 & 0.5044 & 0.7923 & 0.5233 & 0.5432 & 0.8323 & 0.4752 & 0.4934 & 0.7723 \\
      Local-LSTM & 0.5307 & 0.4930 & 0.8316 & 0.5707 & 0.5991 & 0.8914 & 0.5660 & 0.5557 & 0.8235 \\
      Local-Transformer & 0.5998 & 0.5783 & 0.8734 & 0.6002 & 0.6207 & 0.8622 & 0.5505 & 0.5349 & 0.8423 \\
      Local-DH-GEM & 0.5960 & 0.5981 & 0.8692 & 0.5979 & 0.5951 & 0.8766 & 0.5496 & 0.5400 & 0.8451 \\
      \midrule
      \multicolumn{10}{c}{\textsc{Federated}}\\
      \midrule
      FedAvg-Transformer & 0.5761 & 0.5739 & 0.8710 & 0.6343 & 0.6322 & 0.8916 & 0.5736 & 0.5687 & 0.8604 \\
      FedProx-Transformer & 0.5811 & 0.5691 & 0.8547 & 0.6401 & 0.6399 & 0.8946 & 0.5842 & 0.5799 & 0.8588 \\
      FedAvg-DH-GEM & 0.6399 & 0.6423 & 0.8874 & 0.6434 & 0.6457 & 0.8987 & 0.6012 & 0.6023 & 0.8673 \\
      FedProx-DH-GEM & 0.6435 & 0.6467 & 0.8997 & 0.6439 & 0.6464 & 0.8924 & 0.5779 & 0.5805 & 0.8584 \\
      \midrule
      \textbf{MPCAC-FL} & \textbf{0.6650} & \textbf{0.6641} & \textbf{0.9095} & \textbf{0.6718} & \textbf{0.6722} & \textbf{0.9112} & \textbf{0.6191} & \textbf{0.6157} & \textbf{0.8844}\\
      \bottomrule
  \end{tabular}}
\end{table*}

\section{Experiments}

\subsection{Experimental Setup}
\label{sec:Imp}

\subsubsection{Metrics.}
In our experiments, the prediction of demand and supply is a multi-class classification task. Therefore, we mainly adopt Accuracy to evaluate the overall performance of models. Besides, we also use the weighted F1 score (Weighted-F1) and area under the receiver operating characteristic (AUROC) for evaluation.

\subsubsection{Baselines.}
We compare our framework with the following baselines, including statistic-based methods, traditional machine learning methods, and deep learning methods.
\textbf{1)~LV}~(Last Value) is a statistical classifier only using the last trend value of talent demand or supply.
\textbf{2)~LR}~(Logistic Regression) is a linear machine learning model.
\textbf{3)~GBDT}~(Gradient Boosting Decision Tree) is an additive model in a forward stage-wise fashion.
\textbf{4)~LSTM}~\cite{hochreiter1997long} (Long Short-Term Memory) is a typical recurrent neural network for time series prediction.
\textbf{5)~Transformer}~\cite{vaswani2017attention} is an attention mechanism-based model which is very popular for modeling various sequence data.
\new{\textbf{6)~DH-GEM}~\cite{guo2022talent} (Dynamic Heterogeneous Graph Enhanced Meta-learning) is the conference version of MPCAC-FL and the state-of-the-art predictive framework for talent demand-supply joint prediction, which we regard as the performance upper bound of the task.

These models are optionally deployed under three optimization schemes, including
\textbf{1)~Global}: we follow~\cite{guo2022talent} to use the usual supervised training scheme on data of all companies.
\textbf{2)~Local}: we conduct supervised training on the data of each client without the server-side aggregation.
\textbf{3)~Federated}: we leverage federated optimization schemes including \texttt{FedAvg} and \texttt{FedProx} to train deep learning models.}

\subsubsection{Implementation details.}

\textbf{MPCAC-FL.}
For hyper-parameters, we choose the number of trend types $|Y|=5$, the minimum length of fine-grained sequences $L_{\min}=12$, the embedding dim of $h_D^t$, \ie ${dim}_t=16$, the embedding dim of graph node representation ${dim}_g=4$, the head number of multi-head attention in the sequential module as $4$, the feed-forward dimension as $16$, the number of layers in the sequential module as $2$ and the output dimension of any other multi-layer perceptron as $4$. We use Adam Optimization with learning rate as $0.01$, learning rate scheduler reducing rate as $0.9$, step as $4$, and weight decay as $10^{-6}$. The DH-GEM is run on the machine with Intel Xeon Gold 6148 @ 2.40GHz, V100 GPU, and 64G memory.

\textbf{Baselines.}
For traditional models, talent demand and supply input lengths of LV, LR, and GDBT are fixed lengths of $5$, and we also set the company and position index as features for LR and GBDT. For LSTM and Transformer, they follow the structure of DSJED and substitute the sequential module as specific encoders, and the DSAJD as two independent $2$-layer multi-layer perceptron. The input and output dimensions of encoders remain consistent with DH-GEM. Specifically,
\textbf{1)~LV} is implemented by a simple one-layer perceptron;
\textbf{2)~LR} is implemented by a linear regression module and the loss is calculated with $L2$ penalty;
\textbf{3)~GBDT} is implemented by a gradient boosting decision tree with $10$ estimators, $0.1$ learning rate;
\textbf{4)~LSTM} substitutes the sequential module of DSJED as two parameter-independent $2$-layer long short-term memory;
\textbf{5)~Transformer} substitutes the sequential module of DSJED as two parameter-independent $2$-layer Transformer encoder with sinusoidal positional encoding;
\new{\textbf{6)~DH-GEM} is implemented as the setting in~\cite{guo2022talent}.}

\new{\subsection{Overall Results}

The results of MPCAC-FL and all baselines on the IT, FIN, and CONS datasets are presented in Table~\ref{tab:overall}. To provide an overall view of the federated prediction performance, we report the average of demand and supply predictions for Accuracy, F1, and AUROC values. We conduct these experiments under three common optimization schemes, namely \textsc{Global}, \textsc{Local}, and \textsc{Federated}, on different backbone models.

The prediction performance of MPCAC-FL is mostly comparable with the non-federated state-of-the-art method on labor market forecasting tasks, \ie DH-GEM, and outperforms all other baselines. Specifically, MPCAC-FL achieves $97.61\%$, $98.93\%$, and $99.37\%$ accuracy of the DH-GEM on three datasets, respectively, with minimal loss in model utility to ensure constrained data privacy. Compared with other federated optimization methods, MPCAC-FL improves at least $3.4\%$, $4.3\%$, and $3.0\%$ accuracy. This improvement is attributed to the personalized federated aggregation strategy and the generalizable optimizing function for data heterogeneity.

Besides, we observe that local training results in significant performance degradation than global training, \eg $12.5\%$ for DH-GEM and $5.44\%$ for Transformer on IT dataset. This discrepancy arises due to limited access to explicit data in local training, leading to inadequate data on the client side and difficulty in extracting macroscopic information from the total data. Fortunately, federated optimization algorithms retrieved cross-client implicit knowledge to increase data sufficiency, with enhanced accuracy by $7.69\%$ for DH-GEM and $6.64\%$ for Transformer on FIN dataset. Consequently, models with federated learning outperformed locally trained models. However, due to the failure to tackle remarkable heterogeneity and the hunger for parameter-intensive models, several federated algorithms perform equivalent or even worse than local training. Our findings demonstrate the need for designing effective federated learning methods to overcome these challenges.

\subsection{Ablation Study}

\begin{figure}[t]
\centering
\subfigure[Accuracy] {\includegraphics[width=.32\linewidth]{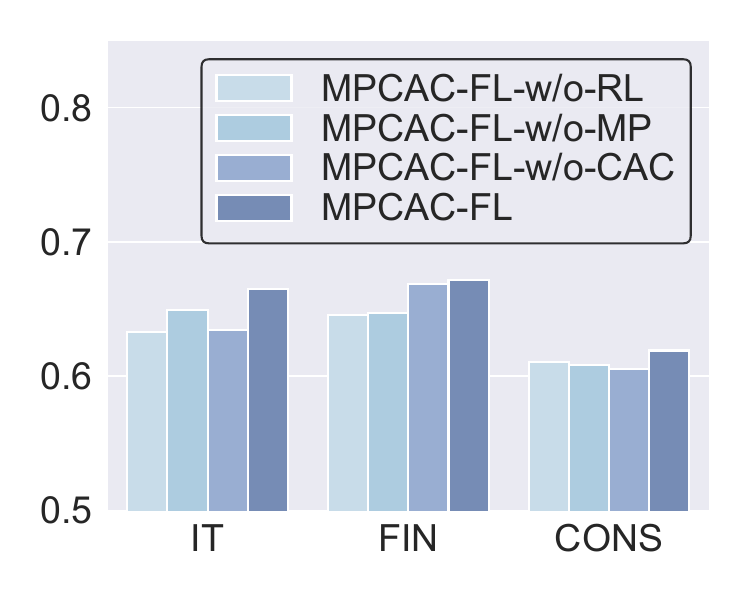}}
\subfigure[Weighted-F1] {\includegraphics[width=.32\linewidth]{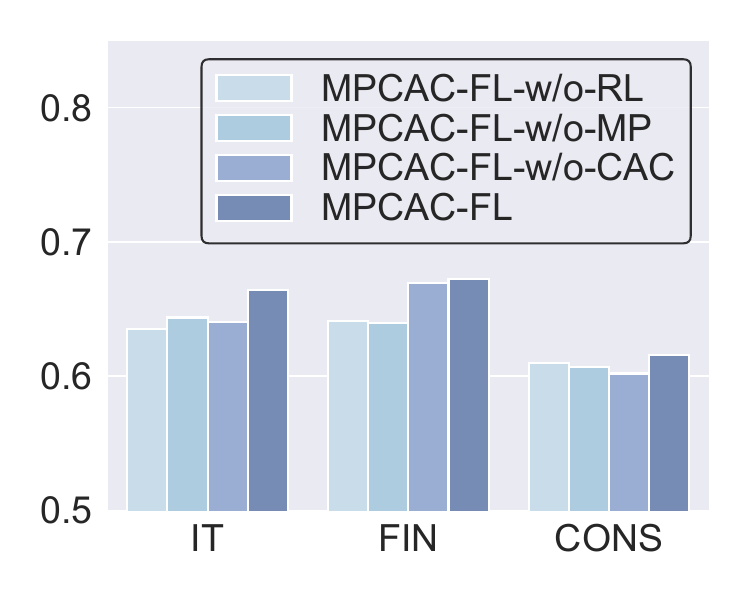}}
\subfigure[AUROC] {\includegraphics[width=.32\linewidth]{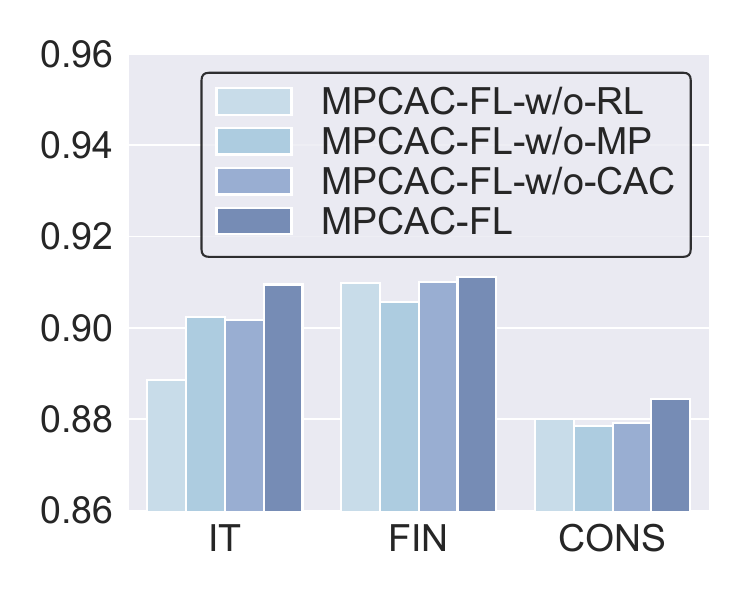}}
\caption{\new{Ablation study on FedLMF task.}}
\label{fig:ablation}
\end{figure}

To evaluate the effectiveness of regularization loss, MAML-enabled personalization, and CAC-FL, we conduct an ablation study with three variant MPCAC-FL algorithms.
\textbf{1)~MPCAC-FL-w/o-RL} is a variant of MPCAC-FL without regularization loss.
\textbf{2)~MPCAC-FL-w/o-MP} is a variant of MPCAC-FL without model-agnostic meta-learning personalization.
\textbf{3)~MPCAC-FL-w/o-CAC} is a variant of MPCAC-FL without convergence-aware clustering.
As shown in Figure~\ref{fig:ablation}, removing one of three modules leads to remarkable performance degradation, which verifies the effectiveness of these modules. Specifically, the accuracy of MPCAC-FL-w/o-RL decreases $4.78\%$ on IT, $3.90\%$ on FIN, and $2.00\%$ on CONS. The most important reason is that regularization loss helps limit the learning space of parameters for better generalization on large-scale clients.
Besides, the accuracy of MPCAC-FL-w/o-MP decreases $2.36\%$ on IT, $3.66\%$ on FIN, and $3.36\%$ on CONS. This result demonstrates that personalization improves federated learning methods with homogeneous assumptions and provides a solution to handle severe data heterogeneity among longtail companies.
Last, the accuracy of MPCAC-FL-w/o-CAC decreases $4.59\%$ on IT, $0.51\%$ on FIN, and $1.44\%$ on CONS. The existing gap can be filled by convergence-aware clustered federated learning due to its separated federated aggregation, which determines that the aggregation only executes among homogeneous clients to avoid the harm brought by heterogeneity.

\subsection{Effectiveness of Addressing the Heterogeneity}

\begin{figure}
  \centering
  \subfigure[Longtail companies performance.]{\includegraphics[width=0.48\linewidth]{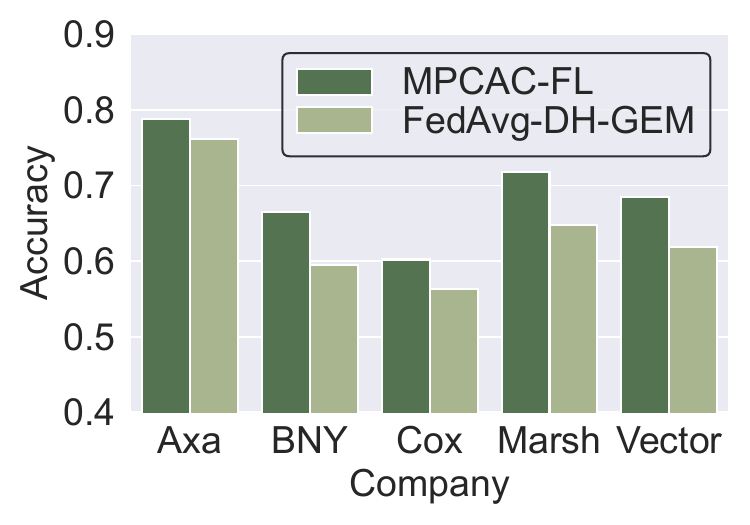}\label{fig:longtail_fin}}
  \subfigure[Training accuracy curve.]{\includegraphics[width=0.48\linewidth]{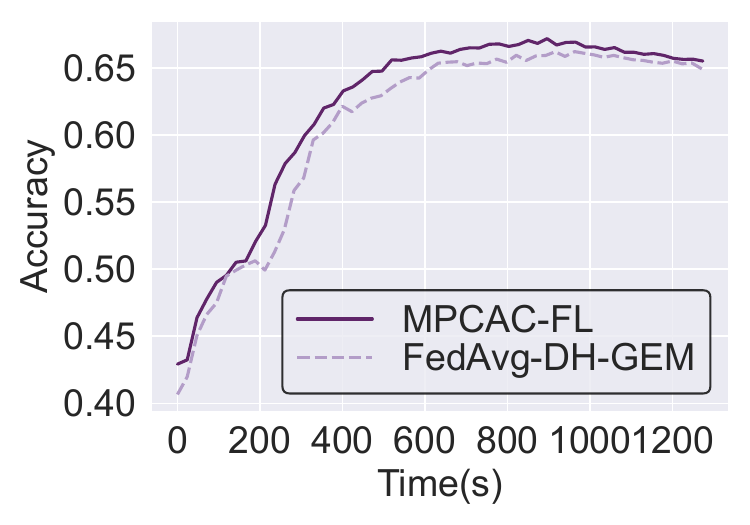}\label{fig:training_fin}}
  \caption{\new{Analysis for addressing the heterogeneity of MPCAC-FL compared with FedAvg-DH-GEM.}}
  \label{fig:analysis_add_hete}
\end{figure}

As investigated in Section~\ref{sec:analysis_longtail}, human resource data is naturally non-IID across companies and mostly obeys longtail distribution. MPCAC-FL is a federated optimization method mainly for heterogeneous clients, whose effectiveness will be shown through the specific comparison of DH-GEM with and without MPCAC-FL (\ie MPCAC-FL and FedAvg-DH-GEM). The prediction improvement statistics are reported in Table~\ref{tab:overall}, where MPCAC-FL performs over $3.9\%$ better than FedAvg-DH-GEM. We also 1)~test accuracy on five long-tail companies, and 2)~draw the training accuracy curve on the FIN validation dataset.
The comparing results are depicted in Figure~\ref{fig:analysis_add_hete}. MPCAC-FL improves remarkable performance on longtail companies (\eg $11.74\%$ for BNY, $10.98\%$ for Marsh, and $10.77\%$ for Vector) shown in Figure~\ref{fig:longtail_fin}, which proves the ability of our approach against data heterogeneity.
In addition, MPCAC-FL achieves higher accuracy and uses less time to reach a similar performance during synchronized training with FedAvg-DH-GEM as shown in Figure~\ref{fig:training_fin}.

\begin{table}[t]
  \caption{Parameter sensitivity of ${dim}_t$ and ${dim}_g$.}
  \label{tab:sense}
  \new{\begin{tabular}{c|ccccc}
      \toprule
      \diagbox{${dim}_t$}{${dim}_g$} & 2 & 4 & 8 & 12 & 16\\
      \midrule
      2 & 0.4816 & 0.4992 & 0.5032 & 0.5129 & 0.5171 \\
      6 & 0.6287 & 0.6142 & 0.6486 & 0.6502 & 0.6578 \\
      12 & 0.6049 & 0.6603 & \textbf{0.6650} & 0.6619 & 0.6595 \\
      20 & 0.5978 & 0.6204 & 0.6540 & 0.6643 & 0.6572 \\
      32 & 0.5204 & 0.6231 & 0.6425 & 0.6575 & 0.6549 \\
      \bottomrule
  \end{tabular}}
\end{table}

\begin{figure}[t]
  \centering
  \subfigure[$L_{\min}$] {\includegraphics[width=0.48\linewidth]{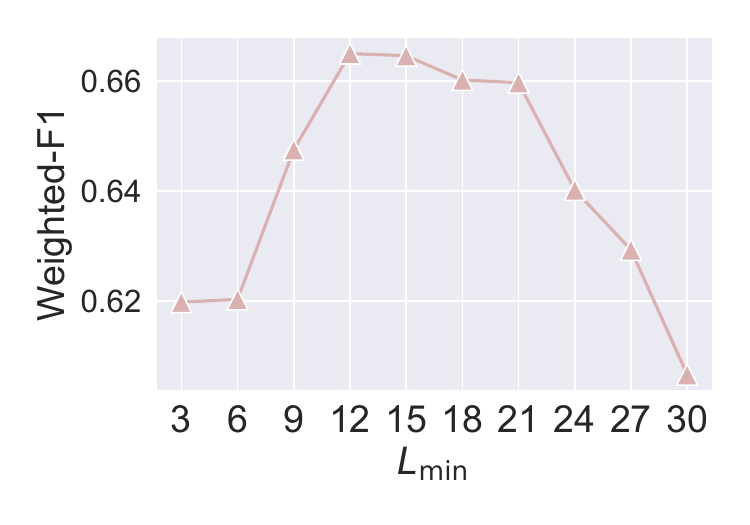} \label{fig:l_sense}}
\subfigure[$|Y|$] {\includegraphics[width=0.48\linewidth]{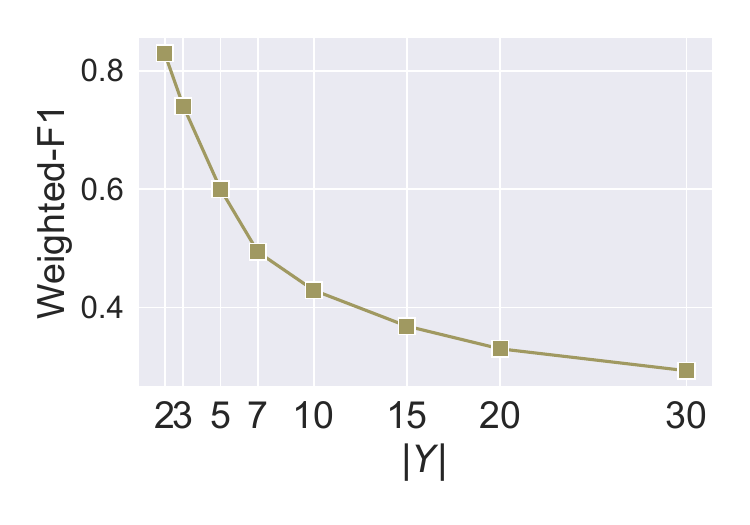} \label{fig:n_sense}}
  \caption{\new{Parameter sensitivity of $L_{\min}$ and $|Y|$.}}
  \label{fig:sense}
\end{figure}

\subsection{Parameters Sensitivity}

Here we first examine the joint sensitivity of temporal embedding dimension ${dim}_t = |h^t|$ and graph node embedding dimension ${dim}_g = |h_c| = |h_p|$ on the IT dataset's accuracy. Our experiments demonstrate that the optimal combination of parameters for MPCAC-FL is ${dim}_t = 12$ and ${dim}_g = 8$, resulting in an accuracy of $0.6650$. The detailed experiment results are shown in Table~\ref{tab:sense}. 
Sequential data provides more information for predicting, requiring a higher dimension than the graph node embedding dimension. However, high embedding dimensions can introduce unnecessary complexity during the model optimization.

Moreover, as shown in Figure~\ref{fig:l_sense}, the input length has a significant impact on model performance. The optimal input sequence length was found to be $L_{\min} = 12$, with both shorter and longer sequences resulting in decreased accuracy. Shorter sequences lack sufficient sequential context, while longer sequences contain more noise that hinders accurate prediction.

Furthermore, we evaluate the sensitivity of the number of trend types $|Y|$ by training with different $|Y|$. As depicted in Figure~\ref{fig:n_sense}, we observe that performance degrades as $|Y|$ increases. This behavior is expected as higher values of $|Y|$ require the model to learn and classify increasingly complex patterns.

\begin{figure}[t]
  \centering
  \includegraphics[width=\linewidth]{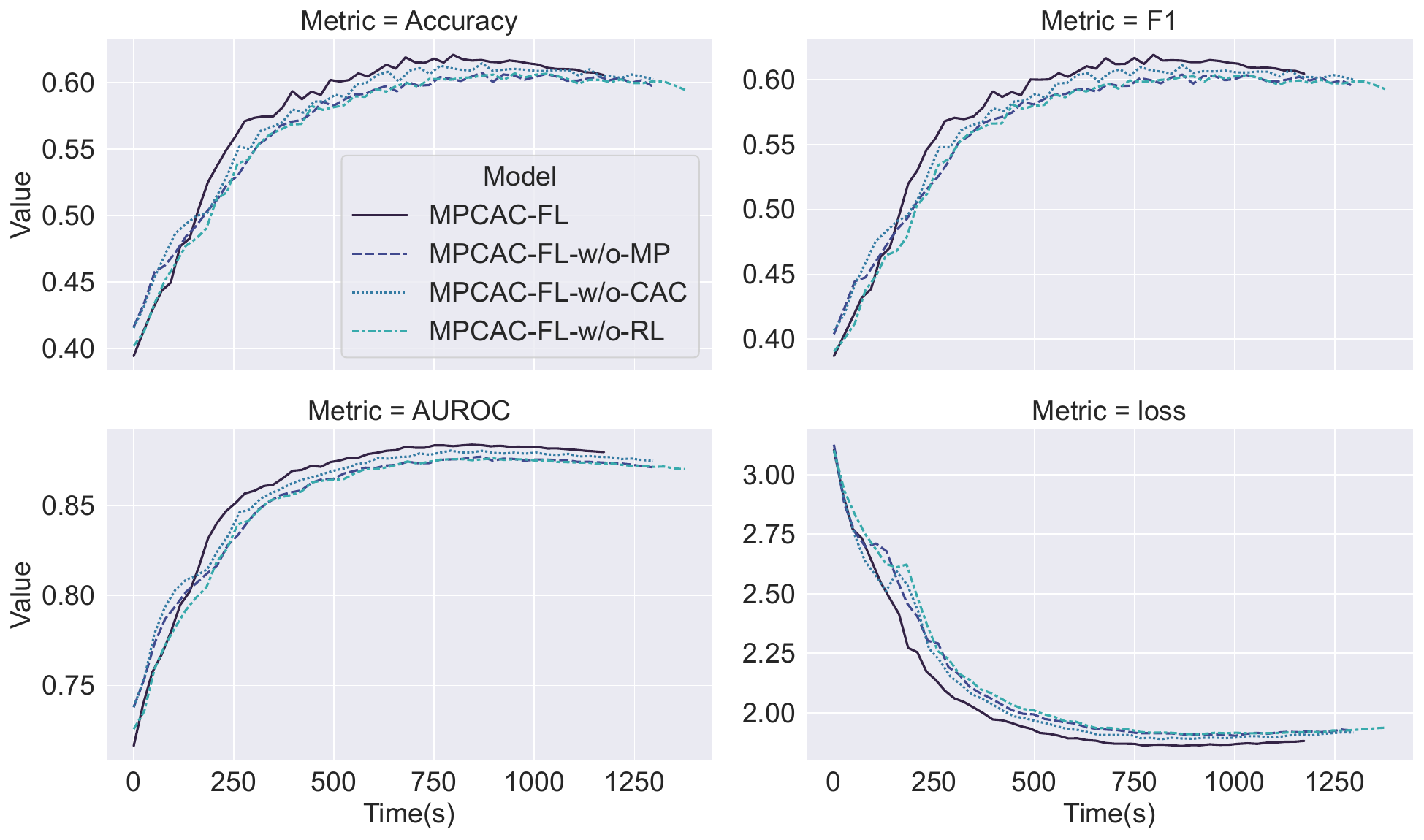}
  \caption{\new{Convergence curves of validations and loss.}}
  \label{fig:convergence}
\end{figure}

\begin{figure*}[t]
    \centering
    \subfigure[Company view.] {\includegraphics[width=0.32\textwidth]{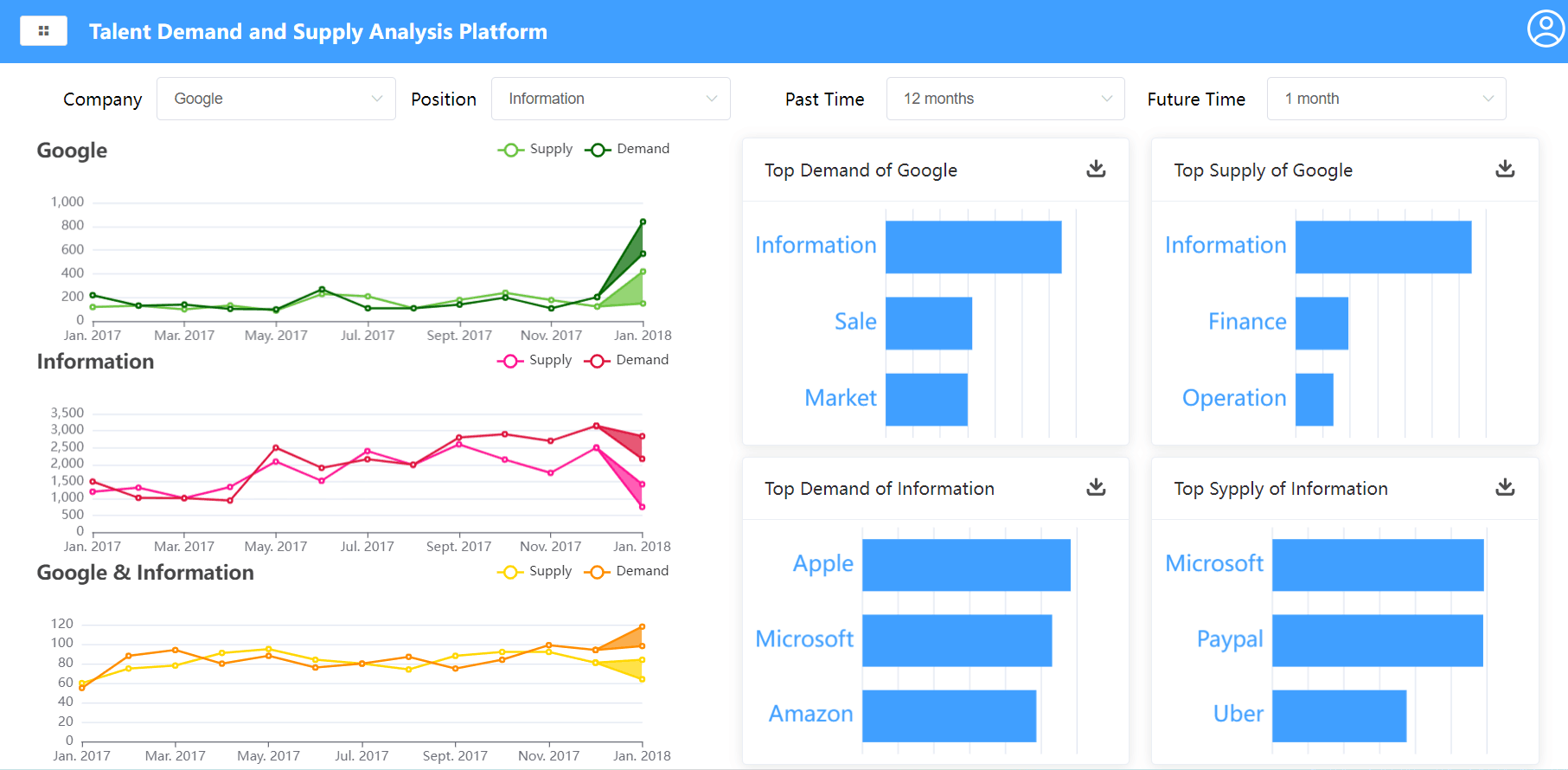}\label{fig:demo_company}}
    \hspace{0.1cm}
    \subfigure[Government view.] {\includegraphics[width=0.32\textwidth]{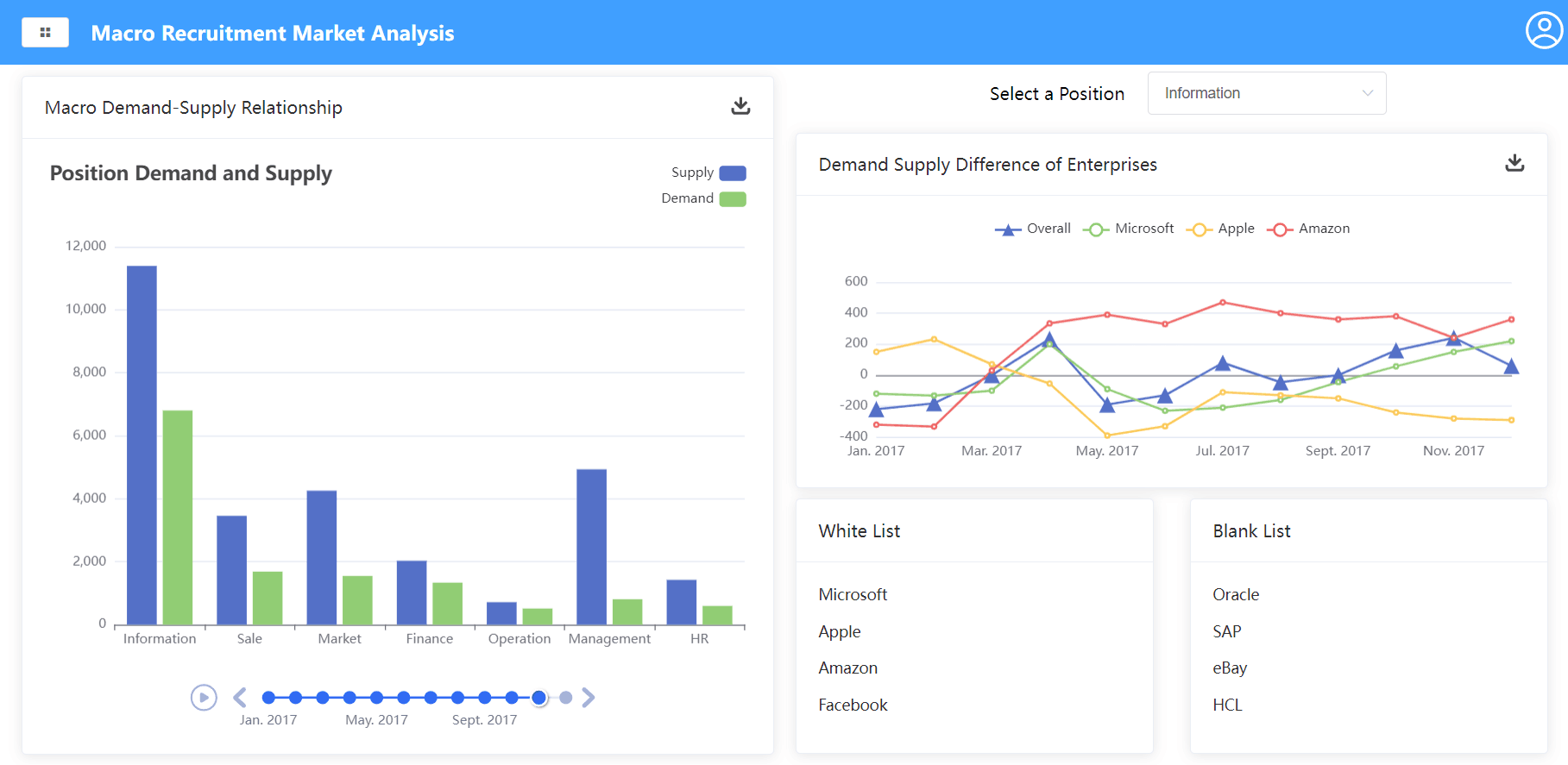}\label{fig:demo_government}}
    \hspace{0.1cm}
    \subfigure[Talent view.] {\includegraphics[width=0.32\textwidth]{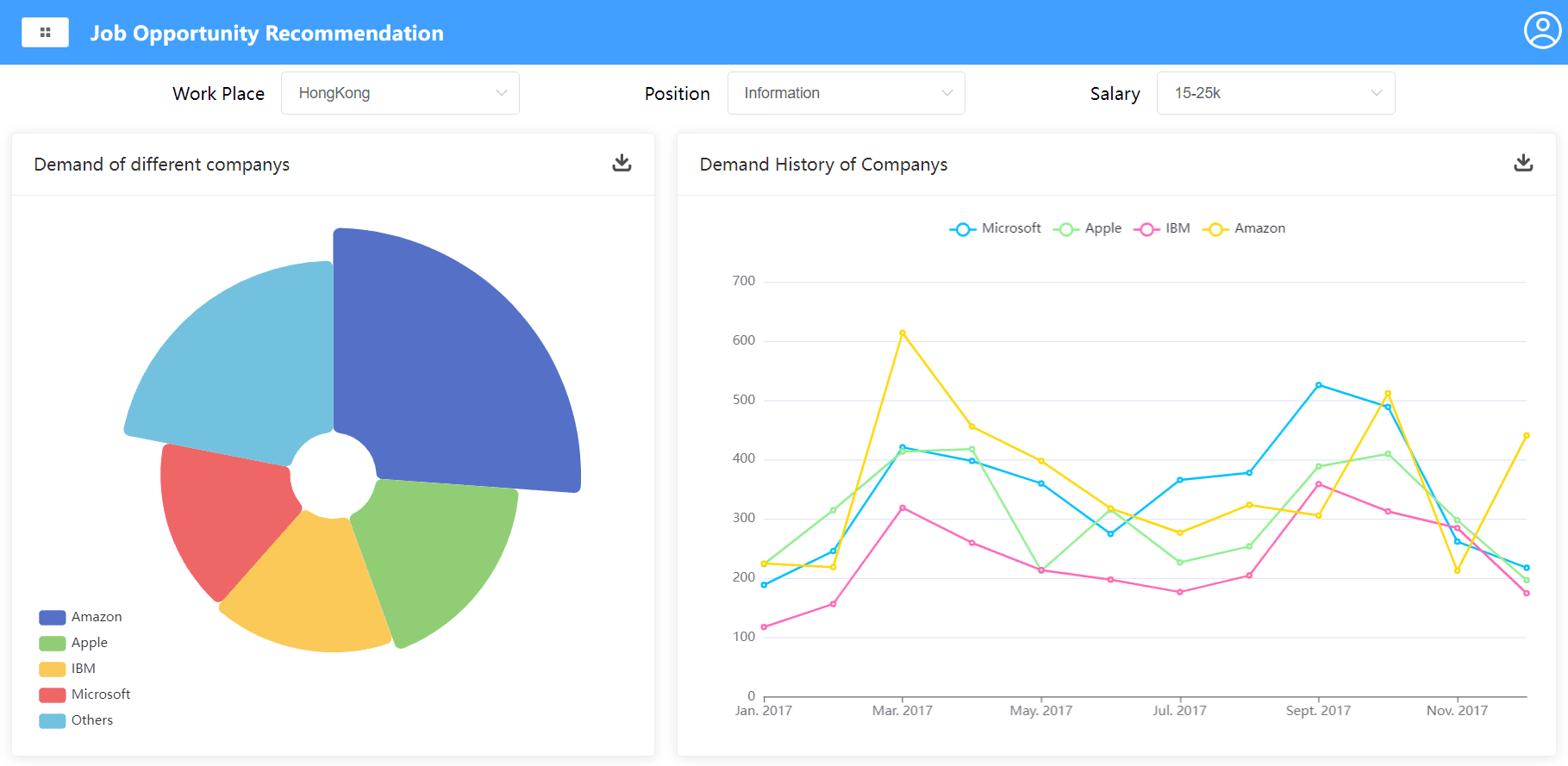}\label{fig:demo_talent}}
    \caption{The screenshot of our deployed system.}
    \label{fig:demo}
\end{figure*}

\subsection{Convergence Analysis}

We analyze the convergence by evaluating three metrics and training loss using the CONS dataset in Figure~\ref{fig:convergence}. We observe that all methods can converge as the training proceeds, and MPCAC-FL converges fastest among competitors since the designed operations target non-IID issues. Three variants perform comparable but worse than MPCAC-FL because of the loss of personalization, distinguishing homogeneous clients, and lack of generalization. Although they have similar performance in the early stage of training, they are stuck after most of the clients locally converge and parameters stay in distant space. MPCAC-FL concerns these factors at the beginning of optimization, which supports its improvement after other models reach their upper bound. Moreover, all federated learning methods will commonly degrade after the peak of accuracy, which further emphasizes the importance of handling heterogeneity for training stability.
}

\subsection{System Deployment}

MPCAC-FL has been deployed in the intelligent human resource system of Baidu. Figure~\ref{fig:demo} shows three views for companies, governments, and talents provided by the system.

\textbf{Company view.} For human resource users in companies, we design the system shown in Figure~\ref{fig:demo_company} that presents both detailed talent demand and supply historic values and the forecasted future trend, and select the top demand or supply of companies or positions with detailed information. These give guidance for employers to grasp the future demand and supply to adjust recruitment strategy and chances to poach talents from supplying companies for our demanding positions.

\new{\textbf{Government view.} Government is concerned about overall demand and supply across industries and monitors company states as shown in Figure~\ref{fig:demo_government}. Specific policies can be introduced to regularize the labor market for disordered positions or companies with abnormal demand and supply.

\textbf{Talent view.} Figure~\ref{fig:demo_talent} depicts the diagram for talents, where for chosen conditions, including workplace, position, and salary, talents can filter potential employers with current and historic quantitative demand. Employees can understand the demand conditioned with their requirements and can adjust their recruitment expectations and choose the most fitted company.}

\section{Related Work}
Overall, the related works of this paper can be summarized into three parts, \ie \emph{labor market trend forecasting}, \emph{federated learning}, \emph{time series prediction}, \emph{graph neural networks}, and \emph{meta learning}.

\subsection{Labor Market Trend Forecasting}
As talent becomes the significant competitiveness between companies, growing attention has been paid to labor trend analysis~\cite{cappelli2008talent}.
For example, a Generalized Least Squares based model was built in~\cite{lin2011factors} by heuristic methods on a governmental realistic data, and analysis on labor market collapse, recovery, and evidence of policy response at COVID-19 onset have been studied~\cite{bartik2020measuring}.
With the technology of machine learning, new emerging methods gradually substitute the traditional ones.
MTLVM~\cite{zhu2016recruitment} is a sequential latent variable model for learning the labor market trend.
Focusing on talent demand forecasting, TDAN~\cite{zhang2021talent} is a data-driven neural sequential approach targeting fine-grained talent demand and its sparsity issue.
NEMO~\cite{li2017nemo} is designed for job mobility, \ie talent supply prediction using contextual embedding.
Ahead~\cite{zhang2021attentive} aims at talent's next career move forecasting with a tailored heterogeneous graph neural network and Dual-GRU.
Fortune Teller~\cite{liu2016fortune} predicts upgrading career paths through fusing information on social networks.
However, these works either focus on talent demand or supply prediction but overlook the intrinsic correlation between talent demand and supply variation.
In addition to demand and supply prediction, other topics of labor market skill validation~\cite{sun2021market, shi2020salience} have been extensively studied in recent years and deployed in human resource systems.

\new{\subsection{Federated Learning}
Federated learning has emerged as a promising solution for collaborative machine learning on decentralized data while preserving data privacy and security~\cite{konevcny2016federated}. It addresses the issues raised by conventional machine learning, in which data is collected in a centralized location and trained using all available data, raising privacy concerns for sensitivity~\cite{bonawitz2017practical}. In contrast, federated learning trains models on data distributed across a large number of participating entities without transferring raw data to a central server~\cite{mcmahan2017communication}.
Federated learning has been widely applied in various business domains. For example, Google has employed federated learning to improve personalized recommendations in Google Play and reduce prediction latency in Google keyboard~\cite{hard2018federated}. In finance, federated learning has been utilized for fraud detection and credit scoring~\cite{chen2020credit}.
In the context of labor market forecasting, federated learning provides a promising solution for training predictive models on decentralized talent demand and supply data while preserving the privacy and confidentiality of individual company data.}

\subsection{Time Series Prediction}
Recently, deep learning-based sequential approaches such as Recurrent Neural Network~(RNN) and Long-Short Term Memory (LSTM)~\cite{hochreiter1997long} have gained unprecedented popularity, due to their capability of learning effective feature representations from complex time series.
Along this line, Transformer~\cite{vaswani2017attention} is a novel Encoder-Decoder architecture solely based on attention mechanisms, which shows a unique superiority in parallel processing and has been adopted as a workhorse for various sequential forecasting tasks.
Meanwhile, talent analysis has been formulated as a time series problem, while sequential models have been applied in labor market trend forecasting. For example, TDAN~\cite{zhang2021talent} leveraged attention based sequential model to forecast talent demand, and Ahead~\cite{zhang2021attentive} used GRU to predict company, position, and duration of the next career move. DH-GEM~\cite{guo2022talent} first jointly modeled talent demand and supply with the incorporation of implicit correlation between two sequences and dynamic labor market information.

\subsection{Graph Neural Network}
Graph Neural Network~(GNN) has been demonstrated as a powerful tool for modeling non-Euclidean relational data structures.
As a basic variant, Graph Convolutional Network~(GCN)~\cite{kipf2016semi} preserves structural proximity for homogeneous graph nodes with a graph convolutional operator.
Besides, Heterogeneous Graph Attention Network~(HAN)~\cite{wang2019heterogeneous} captures sophisticated node-level and semantic-level dependencies for heterogeneous nodes via a hierarchical attention module.
In this work, we propose a dynamic heterogeneous graph neural network to capture the complicated correlation between companies and positions.

\subsection{Meta-learning}
As an emerging learning paradigm, meta-learning has been recognized as a promising way of handling few-shot learning tasks.
As one of the most representative meta-learning approaches, Model-Agnostic Meta-Learning (MAML)~\cite{finn2017model} learns optimal initial parameters of neural networks to transfer globally shared knowledge to new tasks with limited data. As another example, Prototypical network~\cite{snell2017prototypical} achieves better classification accuracy by computing distances between new data and prototype representations.
\new{In this work, we leverage MAML-based meta-learning to personalize heterogeneous companies during federated labor market forecasting model optimization.}

\new{\section{Conclusion}
In this paper, we propose a meta-personalized convergence-aware clustered federated learning to cope with the federated labor market forecasting problem, where an inter-company collaborative model is learned to predict future talent demand and supply trends with privacy preservation. We first design the demand-supply joint encoder-decoder and the dynamic company-position heterogeneous graph convolutional network to implicitly mine shared information between talent demand and supply sequences with the dynamic company and position representations extracted from company-position topological knowledge. Then regularized loss is added to the optimizing objective for improving generalization among data-heterogeneous clients. We also devise the convergence-aware clustered federated learning on the server side to divide clients into groups containing only homogeneous clients, by column-pivoted QR factorization-based spectral clustering. CAC-FL decreases the non-IID issue during federated aggregation for better overall performance. Besides, we select clients using a loss-driven sampler to ensure unfitted local models gain priority to be optimized with knowledge absorbed from others. Extensive experiments on real-world datasets compared with baselines demonstrate that MPCAC-FL achieves the best performance among federated and local methods and predicts comparable with the state-of-the-art model under global training. Importantly, we have deployed MPCAC-FL as a core functional component of the intelligent system of Baidu.}

\bibliographystyle{IEEEtran}
\bibliography{ref}

\begin{thebibliography}{10}
\providecommand{\url}[1]{#1}
\csname url@samestyle\endcsname
\providecommand{\newblock}{\relax}
\providecommand{\bibinfo}[2]{#2}
\providecommand{\BIBentrySTDinterwordspacing}{\spaceskip=0pt\relax}
\providecommand{\BIBentryALTinterwordstretchfactor}{4}
\providecommand{\BIBentryALTinterwordspacing}{\spaceskip=\fontdimen2\font plus
\BIBentryALTinterwordstretchfactor\fontdimen3\font minus \fontdimen4\font\relax}
\providecommand{\BIBforeignlanguage}[2]{{%
\expandafter\ifx\csname l@#1\endcsname\relax
\typeout{** WARNING: IEEEtran.bst: No hyphenation pattern has been}%
\typeout{** loaded for the language `#1'. Using the pattern for}%
\typeout{** the default language instead.}%
\else
\language=\csname l@#1\endcsname
\fi
#2}}
\providecommand{\BIBdecl}{\relax}
\BIBdecl

\bibitem{black2021ai}
J.~S. Black and P.~van Esch, ``Ai-enabled recruiting in the war for talent,'' \emph{Business Horizons}, vol.~64, no.~4, pp. 513--524, 2021.

\bibitem{zhang2021talent}
Q.~Zhang, H.~Zhu, Y.~Sun, H.~Liu, F.~Zhuang, and H.~Xiong, ``Talent demand forecasting with attentive neural sequential model,'' in \emph{Proceedings of the 27th ACM SIGKDD Conference on Knowledge Discovery \& Data Mining}, 2021, pp. 3906--3916.

\bibitem{zhu2016recruitment}
C.~Zhu, H.~Zhu, H.~Xiong, P.~Ding, and F.~Xie, ``Recruitment market trend analysis with sequential latent variable models,'' in \emph{Proceedings of the 22nd ACM SIGKDD international conference on knowledge discovery and data mining}, 2016, pp. 383--392.

\bibitem{li2017nemo}
L.~Li, H.~Jing, H.~Tong, J.~Yang, Q.~He, and B.-C. Chen, ``Nemo: Next career move prediction with contextual embedding,'' in \emph{Proceedings of the 26th International Conference on World Wide Web Companion}, 2017, pp. 505--513.

\bibitem{zhang2021attentive}
L.~Zhang, D.~Zhou, H.~Zhu, T.~Xu, R.~Zha, E.~Chen, and H.~Xiong, ``Attentive heterogeneous graph embedding for job mobility prediction,'' in \emph{Proceedings of the 27th ACM SIGKDD Conference on Knowledge Discovery \& Data Mining}, 2021, pp. 2192--2201.

\bibitem{cappelli2014talent}
P.~Cappelli and J.~Keller, ``Talent management: Conceptual approaches and practical challenges,'' \emph{Annu. Rev. Organ. Psychol. Organ. Behav.}, vol.~1, no.~1, pp. 305--331, 2014.

\bibitem{lin2011factors}
L.~Lin, J.-S. Horng, Y.-C. Chen, and C.-Y. Tsai, ``Factors affecting hotel human resource demand in taiwan,'' \emph{International Journal of Hospitality Management}, vol.~30, no.~2, pp. 312--318, 2011.

\bibitem{azar2022labor}
J.~Azar, I.~Marinescu, and M.~Steinbaum, ``Labor market concentration,'' \emph{Journal of Human Resources}, vol.~57, no.~S, pp. S167--S199, 2022.

\bibitem{sia2013university}
J.~K.~M. Sia, ``University choice: Implications for marketing and positioning,'' \emph{Education}, vol.~3, no.~1, pp. 7--14, 2013.

\bibitem{cappelli2008talent}
P.~Cappelli, ``Talent management for the twenty-first century,'' \emph{Harvard business review}, vol.~86, no.~3, p.~74, 2008.

\bibitem{aviv2001effect}
Y.~Aviv, ``The effect of collaborative forecasting on supply chain performance,'' \emph{Management science}, vol.~47, no.~10, pp. 1326--1343, 2001.

\bibitem{guo2022talent}
Z.~Guo, H.~Liu, L.~Zhang, Q.~Zhang, H.~Zhu, and H.~Xiong, ``Talent demand-supply joint prediction with dynamic heterogeneous graph enhanced meta-learning,'' in \emph{Proceedings of the 28th ACM SIGKDD Conference on Knowledge Discovery and Data Mining}, 2022, pp. 2957--2967.

\bibitem{huang2001effects}
T.-C. Huang, ``The effects of linkage between business and human resource management strategies,'' \emph{Personnel review}, vol.~30, no.~2, pp. 132--151, 2001.

\bibitem{hubbard1998human}
J.~C. Hubbard, K.~A. Forcht, and D.~S. Thomas, ``Human resource information systems: An overview of current ethical and legal issues,'' \emph{Journal of Business Ethics}, pp. 1319--1323, 1998.

\bibitem{mcmahan2017communication}
B.~McMahan, E.~Moore, D.~Ramage, S.~Hampson, and B.~A. y~Arcas, ``Communication-efficient learning of deep networks from decentralized data,'' in \emph{Artificial intelligence and statistics}.\hskip 1em plus 0.5em minus 0.4em\relax PMLR, 2017, pp. 1273--1282.

\bibitem{delaney1996impact}
J.~T. Delaney and M.~A. Huselid, ``The impact of human resource management practices on perceptions of organizational performance,'' \emph{Academy of Management journal}, vol.~39, no.~4, pp. 949--969, 1996.

\bibitem{lepak1999human}
D.~P. Lepak and S.~A. Snell, ``The human resource architecture: Toward a theory of human capital allocation and development,'' \emph{Academy of management review}, vol.~24, no.~1, pp. 31--48, 1999.

\bibitem{delery1996modes}
J.~E. Delery and D.~H. Doty, ``Modes of theorizing in strategic human resource management: Tests of universalistic, contingency, and configurational performance predictions,'' \emph{Academy of management Journal}, vol.~39, no.~4, pp. 802--835, 1996.

\bibitem{li2020deep}
S.~Li, B.~Shi, J.~Yang, J.~Yan, S.~Wang, F.~Chen, and Q.~He, ``Deep job understanding at linkedin,'' in \emph{Proceedings of the 43rd International ACM SIGIR Conference on Research and Development in Information Retrieval}, 2020, pp. 2145--2148.

\bibitem{yao2013learning}
Y.~Yao, R.~Kohli, S.~A. Sherer, and J.~Cederlund, ``Learning curves in collaborative planning, forecasting, and replenishment (cpfr) information systems: An empirical analysis from a mobile phone manufacturer,'' \emph{Journal of Operations Management}, vol.~31, no.~6, pp. 285--297, 2013.

\bibitem{makarius2017addressing}
E.~E. Makarius and M.~Srinivasan, ``Addressing skills mismatch: Utilizing talent supply chain management to enhance collaboration between companies and talent suppliers,'' \emph{Business horizons}, vol.~60, no.~4, pp. 495--505, 2017.

\bibitem{zhu2021federated}
H.~Zhu, J.~Xu, S.~Liu, and Y.~Jin, ``Federated learning on non-iid data: A survey,'' \emph{Neurocomputing}, vol. 465, pp. 371--390, 2021.

\bibitem{vaswani2017attention}
A.~Vaswani, N.~Shazeer, N.~Parmar, J.~Uszkoreit, L.~Jones, A.~N. Gomez, {\L}.~Kaiser, and I.~Polosukhin, ``Attention is all you need,'' in \emph{Advances in neural information processing systems}, 2017, pp. 5998--6008.

\bibitem{boschma2014labour}
R.~Boschma, R.~H. Eriksson, and U.~Lindgren, ``Labour market externalities and regional growth in sweden: The importance of labour mobility between skill-related industries,'' \emph{Regional Studies}, vol.~48, no.~10, pp. 1669--1690, 2014.

\bibitem{guerrero2013employment}
O.~A. Guerrero and R.~L. Axtell, ``Employment growth through labor flow networks,'' \emph{PloS one}, vol.~8, no.~5, p. e60808, 2013.

\bibitem{li2020federated}
T.~Li, A.~K. Sahu, M.~Zaheer, M.~Sanjabi, A.~Talwalkar, and V.~Smith, ``Federated optimization in heterogeneous networks,'' \emph{Proceedings of Machine learning and systems}, vol.~2, pp. 429--450, 2020.

\bibitem{finn2017model}
C.~Finn, P.~Abbeel, and S.~Levine, ``Model-agnostic meta-learning for fast adaptation of deep networks,'' in \emph{International Conference on Machine Learning}.\hskip 1em plus 0.5em minus 0.4em\relax PMLR, 2017, pp. 1126--1135.

\bibitem{damle2019simple}
A.~Damle, V.~Minden, and L.~Ying, ``Simple, direct and efficient multi-way spectral clustering,'' \emph{Information and Inference: A Journal of the IMA}, vol.~8, no.~1, pp. 181--203, 2019.

\bibitem{ng2001spectral}
A.~Ng, M.~Jordan, and Y.~Weiss, ``On spectral clustering: Analysis and an algorithm,'' \emph{Advances in neural information processing systems}, vol.~14, 2001.

\bibitem{brenier1991polar}
Y.~Brenier, ``Polar factorization and monotone rearrangement of vector-valued functions,'' \emph{Communications on pure and applied mathematics}, vol.~44, no.~4, pp. 375--417, 1991.

\bibitem{hochreiter1997long}
S.~Hochreiter and J.~Schmidhuber, ``Long short-term memory,'' \emph{Neural computation}, vol.~9, no.~8, pp. 1735--1780, 1997.

\bibitem{bartik2020measuring}
A.~W. Bartik, M.~Bertrand, F.~Lin, J.~Rothstein, and M.~Unrath, ``Measuring the labor market at the onset of the covid-19 crisis,'' National Bureau of Economic Research, Tech. Rep., 2020.

\bibitem{liu2016fortune}
Y.~Liu, L.~Zhang, L.~Nie, Y.~Yan, and D.~Rosenblum, ``Fortune teller: predicting your career path,'' in \emph{Proceedings of the AAAI conference on artificial intelligence}, vol.~30, no.~1, 2016.

\bibitem{sun2021market}
Y.~Sun, F.~Zhuang, H.~Zhu, Q.~Zhang, Q.~He, and H.~Xiong, ``Market-oriented job skill valuation with cooperative composition neural network,'' \emph{Nature communications}, vol.~12, no.~1, pp. 1--12, 2021.

\bibitem{shi2020salience}
B.~Shi, J.~Yang, F.~Guo, and Q.~He, ``Salience and market-aware skill extraction for job targeting,'' in \emph{Proceedings of the 26th ACM SIGKDD International Conference on Knowledge Discovery \& Data Mining}, 2020, pp. 2871--2879.

\bibitem{konevcny2016federated}
J.~Kone{\v{c}}n{\`y}, H.~B. McMahan, F.~X. Yu, P.~Richt{\'a}rik, A.~T. Suresh, and D.~Bacon, ``Federated learning: Strategies for improving communication efficiency,'' \emph{arXiv preprint arXiv:1610.05492}, 2016.

\bibitem{bonawitz2017practical}
K.~Bonawitz, V.~Ivanov, B.~Kreuter, A.~Marcedone, H.~B. McMahan, S.~Patel, D.~Ramage, A.~Segal, and K.~Seth, ``Practical secure aggregation for privacy-preserving machine learning,'' in \emph{proceedings of the 2017 ACM SIGSAC Conference on Computer and Communications Security}, 2017, pp. 1175--1191.

\bibitem{hard2018federated}
A.~Hard, A.~V. Rao, A.~Mathews, P.~Ramachandran, and A.~Beutel, ``Federated learning for mobile keyboard prediction,'' \emph{arXiv preprint arXiv:1811.03604}, 2018.

\bibitem{chen2020credit}
K.~Chen, A.~Yadav, A.~Khan, and K.~Zhu, ``Credit fraud detection based on hybrid credit scoring model,'' \emph{Procedia Computer Science}, vol. 167, pp. 2--8, 2020.

\bibitem{kipf2016semi}
T.~N. Kipf and M.~Welling, ``Semi-supervised classification with graph convolutional networks,'' \emph{arXiv preprint arXiv:1609.02907}, 2016.

\bibitem{wang2019heterogeneous}
X.~Wang, H.~Ji, C.~Shi, B.~Wang, Y.~Ye, P.~Cui, and P.~S. Yu, ``Heterogeneous graph attention network,'' in \emph{The World Wide Web Conference}, 2019, pp. 2022--2032.

\bibitem{snell2017prototypical}
J.~Snell, K.~Swersky, and R.~S. Zemel, ``Prototypical networks for few-shot learning,'' \emph{arXiv preprint arXiv:1703.05175}, 2017.

\end{thebibliography}

\begin{IEEEbiography}
[{\includegraphics[width=1in,height=1.25in,clip,keepaspectratio]{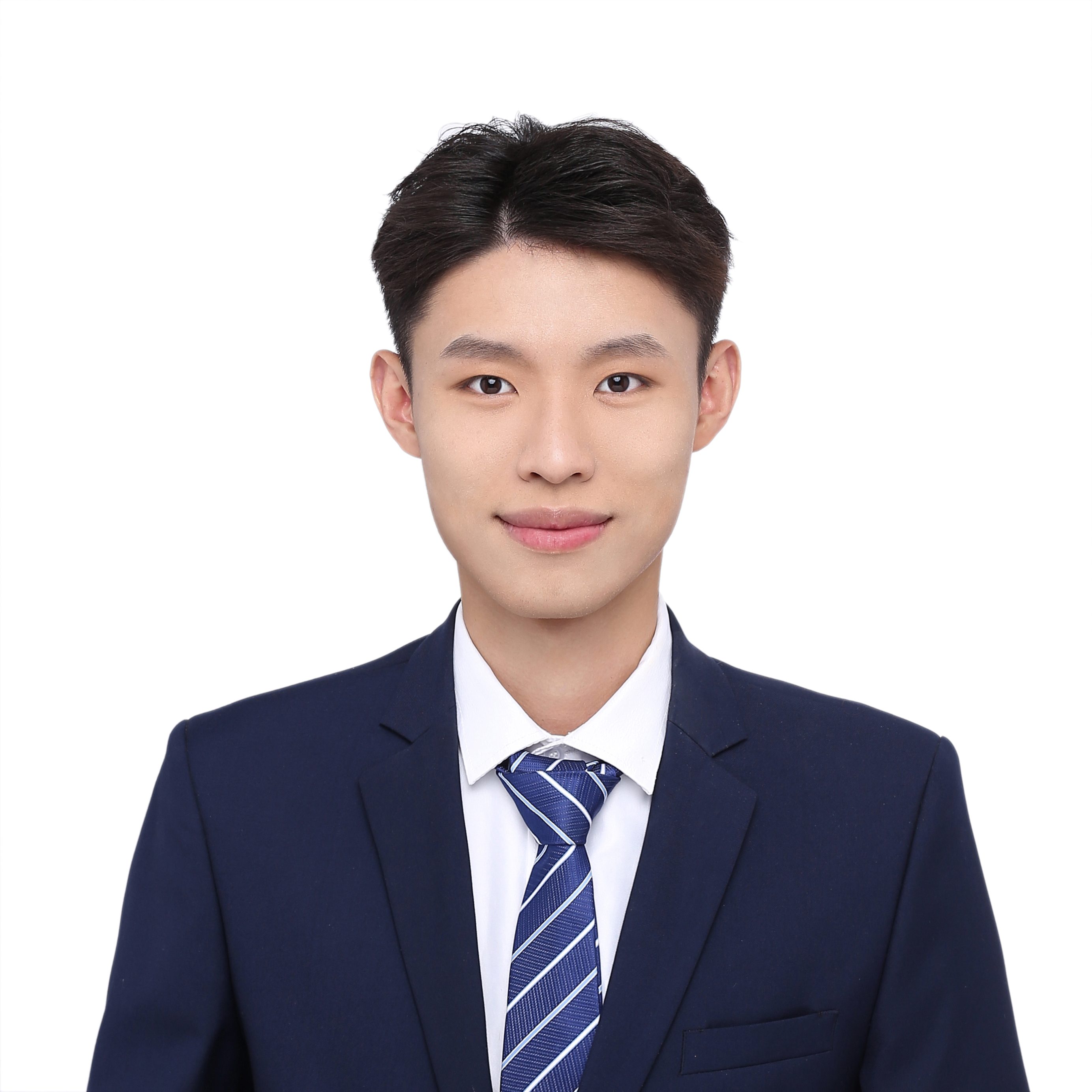}}]{Zhuoning Guo}
is currently pursuing his Ph.D. degree at the Hong Kong University of Science and Technology~(Guangzhou). He received a B.E. degree in Software Engineering from the Harbin Institute of Technology~(HIT) in 2022. His research interests include federated learning, urban computing, and graph learning.  He has published several research papers in prestigious conferences and journals, such as KDD, AAAI, and TVCG.
\end{IEEEbiography}

\begin{IEEEbiography}
[{\includegraphics[width=1in,height=1.25in,clip,keepaspectratio]{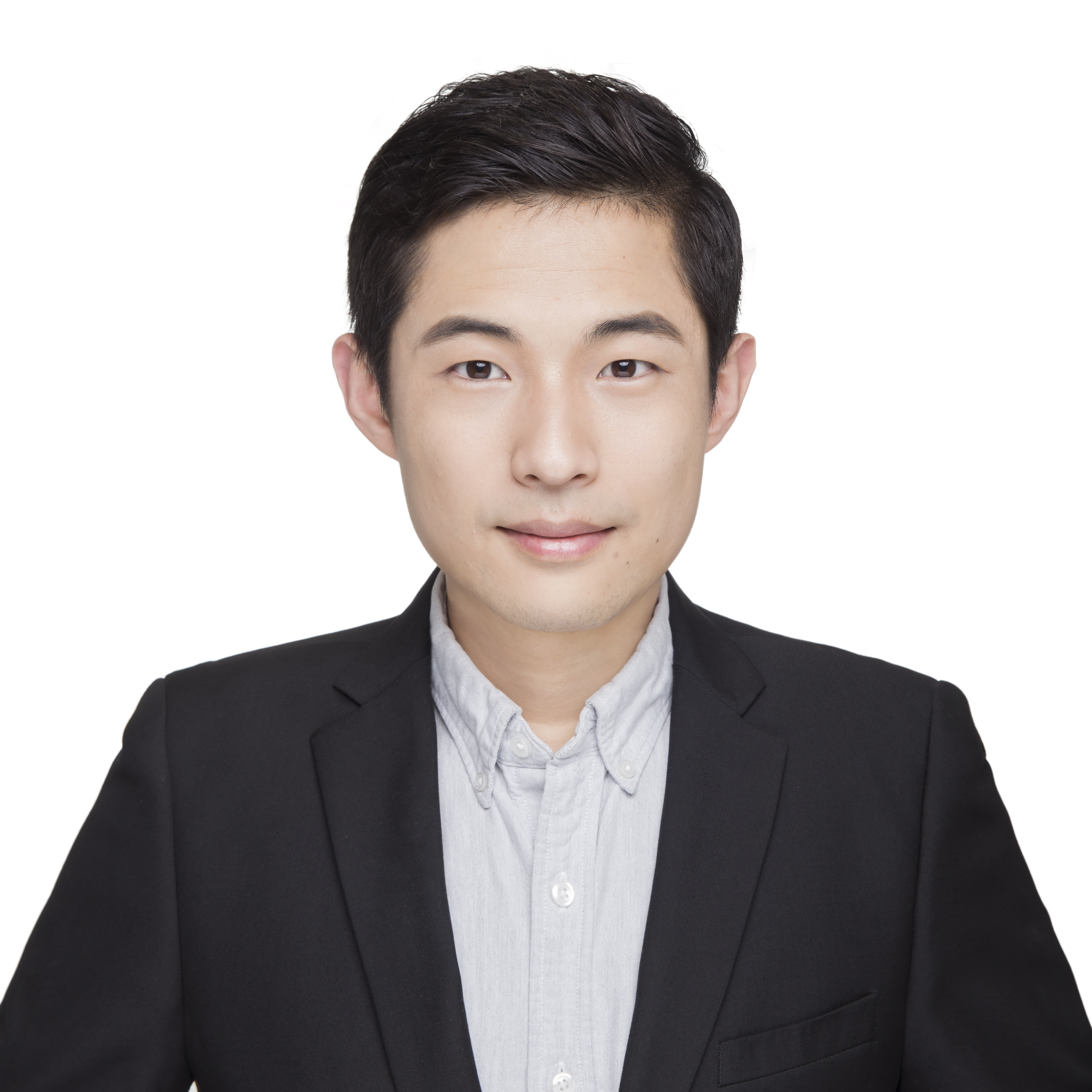}}]{Hao Liu}
is currently an assistant professor at the Thrust of Artificial Intelligence, The Hong Kong University of Science and Technology (Guangzhou). Prior to that, he was a senior research scientist at Baidu Research and a postdoctoral fellow at HKUST. He received the Ph.D. degree from the Hong Kong University of Science and Technology, in 2017 and the B.E. degree from the South China University of Technology (SCUT), in 2012. His general research interests are in data mining, machine learning, and big data management, with a special focus on mobile analytics and urban computing. He has published prolifically in refereed journals and conference proceedings, such as TKDE, KDD, SIGIR, WWW, AAAI, and IJCAI.
\end{IEEEbiography}

\begin{IEEEbiography}
[{\includegraphics[width=1in,height=1.25in,clip,keepaspectratio]{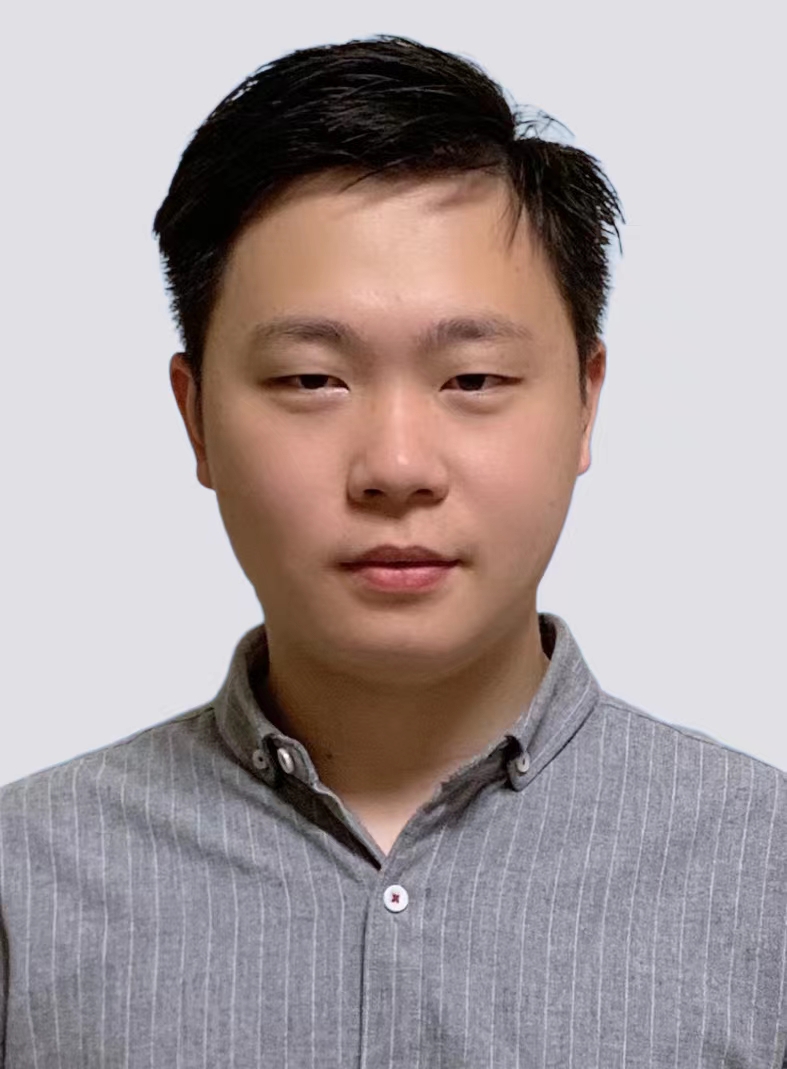}}]
{Le Zhang}
is currently a researcher at Baidu Research. He received his B.E. degree in Software Engineering from Dalian University of Technology in 2016. And he received his Ph.D. degree in Computer Science from University of Science and Technology of China (USTC) in 2022.  His general research interests are data mining and machine learning, with a focus on spatio-temporal modeling and its applications in business intelligence.
\end{IEEEbiography}

\begin{IEEEbiography}
[{\includegraphics[width=1in,height=1.25in,clip,keepaspectratio]{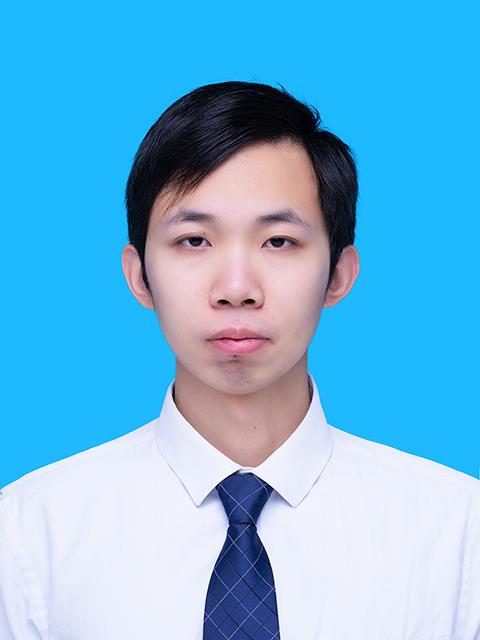}}]{Qi Zhang}
received the Ph.D. degree in Computer Science from the University of Science and Technology of China (USTC), Hefei, China, in 2023. He is currently working as a Postdoctor at the Shanghai Artificial Intelligence Laboratory. He has authored 10 journal and conference papers in the fields of data mining and artificial intelligence, including ACM TOIS, ACM IMWUT, KDD, Nature Communications, ICDM, etc. His research interests include data mining, artificial intelligence, sequential modeling, embodied AI, and natural language processing.
\end{IEEEbiography}

\begin{IEEEbiography}
[{\includegraphics[width=1in,height=1.25in,clip,keepaspectratio]{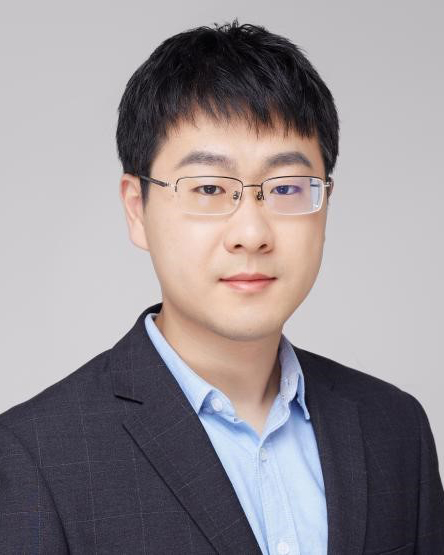}}]{Hengshu Zhu}
received the B.E. and Ph.D. degrees in computer science from the University of Science and Technology of
China (USTC), Hefei, China, in 2009 and 2014, respectively. Previously, he served as the Head of the Baidu Talent Intelligence Center (TIC) and the Principal Scientist of Baidu Research, Baidu Inc., Beijing, China. He is currently the Head of BOSS Zhipin Career Science Lab (CSL). His research interests include data mining and machine learning, with a focus on developing advanced data analysis techniques for innovative business applications. He has published prolifically in refereed journals and conference proceedings and served regularly on the organization and program committees of numerous conferences. He was the recipient of the Distinguished Dissertation Award of CAS (2016), the Distinguished Dissertation Award of CAAI (2016), the Special Prize of President Scholarship for Postgraduate Students of CAS (2014), the Best Student Paper Award of KSEM-2011, WAIM-2013, CCDM-2014, and the Best Paper Nomination of ICDM-2014. He is the Senior Member of IEEE, ACM, and CCF.
\end{IEEEbiography}

\begin{IEEEbiography}[{\includegraphics[width=1in,height=1.25in,clip,keepaspectratio]{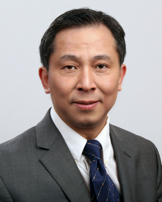}}]{Hui Xiong (F’20) }
is currently a Chair Professor at the Thrust of Artificial Intelligence, The Hong Kong University of Science and Technology (Guangzhou). Dr. Xiong’s research interests include data mining, mobile computing, and their applications in business. Dr. Xiong received his PhD in Computer Science from the University of Minnesota, USA. He has served regularly on the organization and program committees of numerous conferences, including as a Program Co-Chair of the Industrial and Government Track for the 18th ACM SIGKDD International Conference on Knowledge Discovery and Data Mining (KDD), a Program Co-Chair for the IEEE 2013 International Conference on Data Mining (ICDM), a General Co-Chair for the 2015 IEEE International Conference on Data Mining (ICDM), and a Program Co-Chair of the Research Track for the 2018 ACM SIGKDD International Conference on Knowledge Discovery and Data Mining. He received the 2021 AAAI Best Paper Award and the 2011 IEEE ICDM Best Research Paper award. For his outstanding contributions to data mining and mobile computing, he was elected an AAAS Fellow and an IEEE Fellow in 2020.
\end{IEEEbiography}

\vfill

\end{document}